\documentclass[11pt]{article}

\usepackage{amsmath}
\usepackage{amsthm}
\usepackage{amssymb}
\usepackage{mathtools}
\usepackage[cal=euler,scr=boondoxupr,bb=bboldxLight]{mathalpha}
\usepackage{upgreek}
\usepackage{mleftright}
\usepackage[sf,bf]{titlesec}
\usepackage{bm}
\usepackage{euscript}
\usepackage{url}            
\usepackage{booktabs}       
\usepackage{amsfonts}       
\usepackage{nicefrac}       
\usepackage{microtype}      
\usepackage{color}
\usepackage{tabu}
\usepackage{multirow}
\usepackage{float}
\usepackage{booktabs}
\usepackage{algorithm}
\usepackage{algorithmicx}
\usepackage[noEnd=false,indLines=true]{algpseudocodex}
\usepackage{color}
\usepackage{tabu}
\usepackage{comment}
\usepackage{graphicx}
\usepackage{epstopdf}
\usepackage{subcaption}
\usepackage{framed}
\usepackage{enumitem}
\usepackage{textcomp}
\usepackage{setspace}
\usepackage[pdftex]{hyperref}
\usepackage{cleveref}
\usepackage{latexsym}
\usepackage{tcolorbox}
\usepackage[round,compress]{natbib}
\usepackage{alltt}
\usepackage{stmaryrd}
\usepackage{xspace}
\usepackage{euscript}
\usepackage{xfrac}    
\usepackage{nicefrac} 
\usepackage{tikz}
\usepackage{tikz-qtree}
\usepackage{diagbox}
\usepackage{wrapfig}
\usepackage[normalem]{ulem}
\usepackage{todonotes}
\usepackage{inconsolata} 

\usepackage[margin=1in]{geometry}
\usepackage{fancyhdr}
\fancypagestyle{plain}{\fancyhf{} 
	
	\fancyfoot[C]{\textsf{\thepage}}
}
\definecolor{mydarkblue}{rgb}{0,0.08,0.45}
\hypersetup{ %
	colorlinks=true,
	linkcolor=mydarkblue,
	citecolor=mydarkblue,
	filecolor=mydarkblue,
	urlcolor=mydarkblue}

\newcommand{\calB}{\mathcal{B}}

\newcommand{\calD}{\mathcal{D}}

\newcommand{\calF}{\mathcal{F}}

\newcommand{\calZ}{\mathcal{Z}}

\newcommand{\scrO}{\mathscr{O}}

\newcommand{\Var}{\mathrm{Var}}

\newcommand{\Ex}{\mathbb{E}}
\newcommand{\Prob}{\mathbb{P}}

\newcommand{\RR}{\mathbb{R}}

\newcommand{\Rp}{\RR_+}
\newcommand{\Rpp}{\RR_{++}}

\newcommand{\NN}{\mathbb{N}}

\newcommand{\diff}{\mathrm{d}}

\DeclareMathOperator*{\minimize}{minimize}

\newcommand{\sumK}{\sum_{k=1}^K}

\newcommand{\sumn}{\sum_{i=1}^n}

\newcommand{\sfP}{\mathsf{P}}

\newcommand{\sfT}{\mathsf{T}}

\newcommand{\sfV}{\mathsf{V}}

\renewcommand{\mid}{\,|\,}
\newcommand{\midd}{\,|\kern-0.25ex|\,}
\newcommand{\setn}{\llbracket n\rrbracket}

\newcommand{\setd}{\llbracket d\rrbracket}

\newcommand{\setK}{\llbracket K\rrbracket}

\newcommand{\dotp}[2]{\langle #1, #2\rangle}

\newcommand{\RPP}{\ensuremath{\left(0,\infty\right)}}
\newcommand{\RP}{\ensuremath{\left[0,\infty\right)}}

\newcommand{\tvarphi}{\widetilde{\varphi}}

\newcommand{\half}{\sfrac12}

\DeclareMathAlphabet\rsfscr{U}{rsfso}{m}{n}

\let\le\leqslant
\let\ge\geqslant

\let\hat\widehat
\let\tilde\widetilde

\makeatletter
\DeclareFontFamily{OMX}{MnSymbolE}{}
\DeclareSymbolFont{MnLargeSymbols}{OMX}{MnSymbolE}{m}{n}
\SetSymbolFont{MnLargeSymbols}{bold}{OMX}{MnSymbolE}{b}{n}
\DeclareFontShape{OMX}{MnSymbolE}{m}{n}{
	<-6>  MnSymbolE5
	<6-7>  MnSymbolE6
	<7-8>  MnSymbolE7
	<8-9>  MnSymbolE8
	<9-10> MnSymbolE9
	<10-12> MnSymbolE10
	<12->   MnSymbolE12
}{}
\DeclareFontShape{OMX}{MnSymbolE}{b}{n}{
	<-6>  MnSymbolE-Bold5
	<6-7>  MnSymbolE-Bold6
	<7-8>  MnSymbolE-Bold7
	<8-9>  MnSymbolE-Bold8
	<9-10> MnSymbolE-Bold9
	<10-12> MnSymbolE-Bold10
	<12->   MnSymbolE-Bold12
}{}

\let\llangle\@undefined
\let\rrangle\@undefined
\DeclareMathDelimiter{\llangle}{\mathopen}%
{MnLargeSymbols}{'164}{MnLargeSymbols}{'164}
\DeclareMathDelimiter{\rrangle}{\mathclose}%
{MnLargeSymbols}{'171}{MnLargeSymbols}{'171}
\makeatother

\let\le\leqslant
\let\ge\geqslant

\let\hat\widehat
\let\tilde\widetilde

\renewcommand{\left}{\mleft}
\renewcommand{\right}{\mright}

\theoremstyle{plain}
\newtheorem{proposition}{Proposition}
\newtheorem{theorem}[proposition]{Theorem}
\newtheorem{lemma}[proposition]{Lemma}

\theoremstyle{remark}

\theoremstyle{definition}
\newtheorem{assumption}[proposition]{Assumption}
\newtheorem{definition}[proposition]{Definition}

\crefname{assumption}{Assumption}{Assumptions}
\Crefname{assumption}{Assumption}{Assumptions}
\crefname{problem}{Problem}{Problems}
\Crefname{problem}{Problem}{Problems}
\crefname{example}{Example}{Examples}
\Crefname{example}{Example}{Examples}

\newcommand{\lrdotp}[2]{\left\langle #1, #2 \right\rangle}
\newcommand{\norm}[1]{\left\lVert#1\right\rVert}

\newcommand{\matsnorm}[2]{\lvert\kern-0.25ex\lvert\kern-0.25ex\lvert #1 \rvert\kern-0.25ex\rvert\kern-0.25ex\rvert_{#2}}

\newcommand{\Adam}{\textsc{Adam}\xspace}

\newcommand{\SGD}{\textsc{SGD}\xspace}

\newcommand{\AdaGrad}{\textsc{AdaGrad}\xspace}
\newcommand{\AdaGradNorm}{\textsc{AdaGrad-Norm}\xspace}

\newcommand{\AdaDelta}{\textsc{AdaDelta}\xspace}
\newcommand{\ResNet}{\textsc{ResNet}\xspace}

\newcommand{\AdAdaGrad}{\textsc{AdAdaGrad}\xspace}
\newcommand{\AdAdaGradNorm}{\textsc{AdAdaGrad-Norm}\xspace}
\newcommand{\AdaSGD}{\textsc{AdaSGD}\xspace}
\newcommand{\AdAdam}{\textsc{AdAdam}\xspace}

\newcommand{\xkpo}{x_{k+1}}
\newcommand{\xkmo}{x_{k-1}}
\newcommand{\gradFxk}{\nabla F(x_k)}

\newcommand{\vkmo}{v_{k-1}}
\newcommand{\gkmo}{g_{k-1}}
\newcommand{\vkj}{v_{k,j}}
\newcommand{\vkmoj}{v_{k-1,j}}
\newcommand{\gkj}{g_{k,j}}

\newcommand\keywordsname{\sffamily Keywords}
\newcommand\MSCcodesname{\sffamily Mathematics Subject Classification (2020)}
\newenvironment{@abssec}[1]{%
     \if@twocolumn
       \section*{#1}%
     \else
       \vspace{.05in}\footnotesize
       \parindent .2in
         {\upshape\bfseries #1 }\ignorespaces 
     \fi}
     {\if@twocolumn\else\par\vspace{.1in}\fi}

\newenvironment{@doisec}[1]{%
     \if@twocolumn
       \section*{#1}%
     \else
       \vspace{.05in}\footnotesize
       \parindent .2in
         {\upshape\bfseries #1. }\ignorespaces\ignorespaces
     \fi}
     {\if@twocolumn\else\par\vspace{.1in}\fi}

\newenvironment{keywords}{\begin{@abssec}{\keywordsname}}{\end{@abssec}}
\newenvironment{MSCcodes}{\begin{@abssec}{\MSCcodesname}}{\end{@abssec}}

\begin{document}
    \title{\sffamily \AdAdaGrad: \\Adaptive Batch Size Schemes for Adaptive Gradient Methods}
    \author{
        Tim Tsz-Kit Lau%
        \thanks{Department of Biostatistics, Epidemiology and Informatics, University of Pennsylvania, Philadelphia, PA 19104, USA; Email: \href{mailto:timlautk@u.northwestern.edu}{\texttt{timlautk@upenn.edu}}. 
        }      
        \and Han Liu%
        \thanks{Department of Computer Science and Department of Statistics and Data Science, Northwestern University, Evanston, IL 60208, USA; Email: \href{mailto:hanliu@northwestern.edu}{\texttt{hanliu@northwestern.edu}}. }
        \and Mladen Kolar%
        \thanks{Department of Data Sciences and Operations, University of Southern California Marshall School of Business, Los Angeles, CA 90089, USA; Email: \href{mailto:mkolar@marshall.usc.edu}{\texttt{mkolar@marshall.usc.edu}}.}
    }
    
    \maketitle

\begin{abstract}%
    The choice of batch size in minibatch stochastic gradient optimization is critical for both optimization and generalization performance in large-scale model training. Although large-batch training is arguably the dominant paradigm in large-scale deep learning because of hardware advances, model generalization often deteriorates relative to small-batch training, leading to the so-called ``generalization gap.'' To mitigate this issue, we investigate adaptive batch size strategies derived from adaptive sampling methods, which were originally developed for stochastic gradient descent. Given the strong interplay between learning rates and batch sizes, together with the prevalence of adaptive gradient methods in deep learning, we emphasize the need for adaptive batch size strategies in these settings. We introduce \AdAdaGrad and its scalar variant \AdAdaGradNorm, which progressively increase batch sizes during training while performing updates with \AdaGrad and \AdaGradNorm, respectively. We prove that \AdAdaGradNorm converges with high probability at a rate of $\mathscr{O}(1/K)$ to a first-order stationary point of a smooth nonconvex function within $K$ iterations. \AdAdaGrad also exhibits similar convergence properties when combined with a novel coordinate-wise variant of our adaptive batch size strategy. We corroborate our theoretical claims with image-classification experiments that highlight the merits of the proposed schemes in terms of both training efficiency and model generalization. Our work highlights the potential of adaptive batch size strategies for adaptive gradient optimizers in large-scale model training.
\end{abstract}



\section{Introduction}
Large-scale optimization algorithms \citep{bottou2018optimization} form the foundation of deep learning success in the era of generative AI. Minibatch stochastic gradient descent (\SGD) \citep{robbins1951} and its many variants, together with batch-sampling techniques, are the main workhorses for training deep neural networks. However, training deep neural networks, such as those used in transformers, is notoriously challenging because of their high dimensionality and nonconvex loss landscapes. This complexity necessitates extensive hyperparameter tuning and sophisticated training strategies to avoid premature divergence and training instabilities. Consequently, training deep learning models often appears more as an art than a science. The most critical hyperparameter is arguably the learning rate (or step size). Adaptive gradient methods with adaptive learning rates, such as \AdaDelta \citep{zeiler2012adadelta}, \AdaGrad \citep{duchi2011adagrad}, and \Adam \citep{kingma2015}, are now prevalent because they reduce the need for meticulous tuning and complex learning-rate schedules, which are typically required for \SGD. Another important yet frequently overlooked hyperparameter is the batch size. It governs the trade-off between computational efficiency and model generalization by controlling the magnitude of noise in batch gradients. However, batch-size selection in deep learning remains largely heuristic, such as using a constant batch size for convolutional networks or a linear warmup for large language models \citep{brown2020language,rae2021scaling,hoffmann2022training}, and is usually predetermined before training begins. Furthermore, from a hardware-utilization perspective, using a large number of distributed computational resources (i.e., GPUs or TPUs) often necessitates large-batch training when parallel minibatch \SGD \citep{zinkevich2010parallelized,dean2012large} is employed. In addition, the intricate relationship between learning rates and batch sizes deserves attention. Specifically, \citet{smith2018dont} showed an equivalence between reducing step sizes and increasing batch sizes, but this principle applies mainly to \SGD. The impact of varying batch sizes in adaptive gradient methods has not yet been fully explored.

In light of this, our work seeks to clarify how to determine suitable batch sizes for adaptive gradient methods. We aim to introduce automated schedules that dynamically decide when to increase the batch size, and by how much, based on training needs. Our approach is theoretically grounded and relies on the statistics of batch gradients at the iterates, thereby adapting to training dynamics. The proposed schedules can reduce the generalization gap in large-batch training while still making use of large batches in later stages of training. Our strategies are based on adaptive sampling methods \citep{byrd2012sample,bollapragada2018adaptive}. In the context of deep learning, \citet{de2016big,de2017automated} numerically demonstrated the effectiveness of these methods when combined with \AdaDelta. However, the convergence properties of such adaptive sampling methods for adaptive gradient methods have not been thoroughly investigated, leaving a gap between theory and practice. Moreover, in existing adaptive sampling methods developed mainly for \SGD, step sizes are often fixed or adjusted using backtracking line-search procedures (mainly for convex problems). This becomes computationally impractical or inefficient for large models, especially given the nonconvex nature of deep neural networks.

The development of adaptive batch size strategies for deep learning is not new; examples include Big Batch \SGD \citep{de2016big,de2017automated}, CABS \citep{balles2017coupling}, AdaBatch \citep{devarakonda2017adabatch}, SimiGrad \citep{qin2021simigrad}, and AdaScale \SGD \citep{johnson2020adascale}, which adapts learning rates for large batches rather than directly adjusting batch sizes. However, these methods often lack a principled basis with rigorous convergence guarantees, or they are limited to analyses of \SGD under restrictive conditions (e.g., convexity or the Polyak--\L{}ojasiewicz condition), despite the prevalence of adaptive gradient methods in deep learning. Moreover, these approaches still require choices about learning-rate adjustment, such as line-search routines or schedulers, leaving a gap in full adaptivity. In addition, strategies for determining new batch sizes may rely on heuristic rules, such as geometric growth or decay rates \citep{qin2021simigrad}.

\subsection{Contributions}
In this work, our objective is to bridge the gap between theory and practice for adaptive batch size schemes used with adaptive gradient methods. We introduce two main adaptive batch size schemes, grounded in the so-called \emph{adaptive sampling methods} \citep{byrd2012sample,friedlander2012hybrid,bollapragada2018adaptive}, tailored to adaptive gradient methods. Our focus is on \AdaGrad and its norm variant \AdaGradNorm, which are among the simplest and most extensively studied adaptive gradient methods. Developing adaptive batch size schemes for these methods is therefore of both theoretical and practical interest.

The technical contributions of this work are threefold.
From a theoretical perspective, we establish a sublinear convergence rate (with high probability) for our proposed methods when combined with \AdaGradNorm and \AdaGrad on smooth nonconvex objectives. This substantially broadens the existing body of work, which has mainly focused on \SGD. Moreover, we relax the Lipschitz smoothness condition on the objective function by adopting the generalized smoothness concept introduced in \citet{zhang2020why,zhang2020improved}. This adaptation allows for more general and realistic applications in contemporary deep learning practice. On the empirical side, we demonstrate the effectiveness of our proposed methods through a range of numerical experiments on image-classification tasks. These experiments highlight the benefits of the adaptivity of our schemes, driven by both adaptive batch sizes and adaptive step sizes. Finally, we provide an efficient implementation of the proposed approach in PyTorch, using the \texttt{torch.func} module for efficient parallel computation of per-sample gradients via the vectorized map function \texttt{vmap}. 
An open-source implementation is available in the GitHub repository: \url{https://github.com/timlautk/AdAdaGrad}. 
To the best of our knowledge, our proposed methods are the first adaptive batch size schemes based on adaptive sampling methods for adaptive gradient methods that come with convergence guarantees, strong empirical performance, and efficient implementations in deep learning libraries.

\paragraph{Theory--implementation distinction.}
The convergence results in this paper are stated for an idealized exact-test version of the proposed adaptive batch-size schemes. In this version, the batch size at iteration $k$ is assumed to be chosen so that the corresponding population-level variance condition holds at $x_k$. The practical implementation, by contrast, uses empirical batch statistics computed from the current minibatch to select the next batch size. Thus, the theory establishes convergence for the population-level adaptive-sampling principle underlying the method, while the experiments evaluate a computationally feasible approximate implementation. We make this distinction explicit throughout the paper.

\section{Related Work}
\label{sec:related}
We provide an overview of related work on batch sizes in model training and adaptive sampling methods, as well as on the convergence of stochastic gradient methods.

\subsection{Large-Batch Training}
In stochastic gradient optimizers, batch sizes play a crucial role in controlling the variance of stochastic gradients as estimators of full deterministic gradients. Although the noise induced by stochastic gradients can be beneficial in nonconvex optimization, the trend toward larger batches in large-scale model training has become standard thanks to advances in parallel hardware such as GPUs, which substantially reduce training time. The notion of large-batch training in deep learning was popularized in \citet{smith2018bayesian,smith2018dont} and has been widely adopted in applications such as ImageNet classification \citep{goyal2017accurate} and BERT training \citep{you2020large}. Since \citet{lecun2002efficient}, it has been recognized that model generalization can deteriorate in large-batch training. Large-batch training tends to yield loss landscapes with many sharp minima, which are harder to escape during optimization and hence can lead to worse generalization \citep{keskar2017on}. The impact of batch sizes has been examined more systematically in later work. For example, \citet{mccandlish2018empirical} introduced an empirical model for large-batch training without rigorous theoretical proof, postulating the existence of critical batch sizes through extensive numerical simulations on convolutional neural networks (CNNs), LSTMs, and VAEs. Meanwhile, \citet{kaplan2020scaling} focused on transformers. \citet{zhang2019algorithmic} explored how critical batch sizes vary with optimizer characteristics, including momentum, preconditioning, and exponential moving averages, through both large-scale experiments and a noisy quadratic model. \citet{granziol2022learning} applied random matrix theory to examine the batch Hessian, theoretically deriving learning-rate scaling rules as a function of batch size for both \SGD (\emph{linear}) and adaptive gradient methods (\emph{square-root}), with experimental validation. Although the ``generalization gap'' can be narrowed by using larger learning rates proportional to batch size to maintain the gradient-noise scale \citep{hoffer2017train,smith2018bayesian,smith2018dont}, it cannot be completely eliminated. However, \citet{shallue2019measuring} investigated the impact of batch sizes in the context of data parallelism and empirically characterized the effects of large-batch training, finding no evidence of degraded generalization performance.

\subsection{Adaptive Sampling Methods}
In stochastic optimization, a more theoretically grounded approach known as \emph{adaptive sampling methods} has been developed for batch (or minibatch) algorithms. \citet{byrd2012sample} introduced a method called the norm test, which adaptively increases the batch size throughout the optimization process. The rationale behind the norm test traces back to \citet{carter1991global} and represents a more general condition than the one proposed by \citet{friedlander2012hybrid}. Both approaches establish linear convergence when the batch size increases geometrically. \citet{bollapragada2018adaptive} proposed the augmented inner product test, which allows for more gradual increases in batch size than the norm test. Furthermore, \citet{cartis2018global} introduced a relaxation of the norm test that allows its condition to be violated with probability less than $0.5$. This family of adaptive sampling methods belongs to the class of variance-reduced optimization algorithms \citep{johnson2013accelerating,reddi2016stochastic} widely used in machine learning \citep{gower2020variance}.

Adaptive sampling methods have also been extended to problems beyond unconstrained optimization. \citet{xie2023constrained} proposed proximal extensions of the norm test and the inner product test for minimizing a convex composite objective consisting of a stochastic function and a deterministic, potentially nonsmooth function. \citet{beiser2023adaptive} studied deterministic constrained problems, including cases with nonconvex objectives. Moreover, adaptive sampling methods have been applied to a range of optimization problems and algorithms, including \citet{bollapragada2018progressive} for L-BFGS, \citet{berahas2022adaptive} for sequential quadratic programming (SQP) in equality-constrained problems, \citet{bollapragada2023adaptive} for augmented Lagrangian methods, and \citet{bollapragada2023newton} for quasi-Newton methods.

\subsection{Convergence Results of Stochastic Gradient Methods}
We provide a brief overview of convergence results for \SGD and adaptive gradient methods.
The convergence of \SGD for smooth nonconvex functions was first analyzed in \citet{ghadimi2013stochastic}, under the assumption of a uniform bound on the variance of stochastic gradients. \citet{arjevani2023lower} later established a tight lower bound. \citet{bottou2018optimization} extended the convergence result using the so-called \emph{affine variance} noise model. Different assumptions on the second moment of stochastic gradients appearing in the literature were reviewed in \citet{khaled2023better}, with the goal of developing a general convergence theory for \SGD on smooth nonconvex functions.

Another line of work concerns convergence guarantees in expectation for adaptive gradient methods. \citet{ward2019adagrad,ward2020adagrad,li2019convergence,faw2022power} each established a convergence rate of $\tilde{\scrO}(1/\sqrt{K})$ for \AdaGrad under different assumptions on stochastic gradients. \citet{defossez2022a} gave a simple proof of the convergence of both \AdaGrad and a simplified version of \Adam through a unified formulation.

Regarding high-probability convergence bounds, \citet{ghadimi2013stochastic} established high-probability convergence for \SGD with a properly tuned learning rate, assuming known smoothness and sub-Gaussian stochastic-gradient noise bounds. Under the same assumptions, \citet{zhou2018convergence,li2020high} obtained similar results for (delayed) \AdaGrad. Again, under different assumptions on stochastic gradients and using different proof techniques, \citet{kavis2022high,faw2023beyond,wang2023convergence,attia2023sgd,liu2023high} derived high-probability convergence rates for \textsc{AdaGrad}(-\textsc{Norm}).

\section{Problem Formulation}
\label{sec:setting}
In this section, we lay out the general problem formulation considered in this work.

\subsection{Notation}
We define $\setn \coloneqq \{1, \ldots, n\}$ for $n\in\NN^*\coloneqq\NN\setminus\{0\}$, $\Rp\coloneqq\RP$, and $\Rpp\coloneqq\RPP$. We denote the inner product in $\RR^d$ by $\dotp{\cdot}{\cdot}$ and its induced $\ell_2$-norm by $\|\cdot\|$. For a vector $x\in\RR^d$, $[x]_j$ denotes its $j$th coordinate ($j\in\setd$). For a function $f\colon\RR^d\to\RR\cup\{\pm\infty\}$, $\partial_j f$ denotes its partial derivative with respect to the $j$th coordinate, for $j\in\setd$. The ceiling function is denoted by $\lceil\cdot\rceil$.

\subsection{Problem Setting}
We consider the problem of minimizing the \emph{expected risk} $\Ex_{\xi\sim\sfP}[f(x; \xi)]$ with respect to $x\in\RR^d$, where the random variable $\xi \in \calZ\subseteq\RR^p$ is distributed according to the unknown true data distribution $\sfP$. We approximate $\sfP$ by the empirical distribution $\hat{\sfP} = \frac1n\sumn \updelta_{\xi_i}$, where $\{\xi_i\}_{i\in\setn}$ is a sample of size $n\in\NN^*$. This leads to the \emph{empirical risk minimization} problem:
\begin{equation*}
    \minimize_{x\in\RR^d} \ F(x) \coloneqq \Ex_{\xi\sim\hat{\sfP}}[f(x; \xi)] =  \frac1n\sumn f(x; \xi_i).
\end{equation*}
When the sample size $n$ is large, the gradient of $F$ is approximated by its batch counterparts. Given a batch of samples $\calB\subset\setn$ of size $b\coloneqq|\calB|$, we define the batch loss associated with $\calB$ by $F_{\calB}(x) \coloneqq\frac{1}{b}\sum_{i\in\calB} f(x; \xi_i)$. If $f(\cdot; \xi)$ is continuously differentiable for every $\xi\in\calZ$, then the loss and batch-loss gradients are given, respectively, by $\nabla F(x) =  \frac1n\sumn \nabla f(x; \xi_i)$ and $\nabla F_{\calB}(x) =\frac{1}{b}\sum_{i\in\calB} \nabla f(x; \xi_i)$. The batch gradient is an unbiased estimator of the full gradient, that is, $\Ex_{\calB}[\nabla F_{\calB}(x)] = \nabla F(x)$ for every $x\in\RR^d$. In many applications, including deep learning, the objective function $F$ is nonconvex. Thus, we consider the problem of finding a (first-order) $\varepsilon$-stationary point $x^\star\in\RR^d$ satisfying $\|\nabla F(x^\star)\|^2 \le \varepsilon$, rather than a global minimum, which is generally intractable.

\section{Adaptive Sampling Methods}
\label{sec:adaptive_sampling}

The adaptive batch size schemes proposed in this paper are based on a family of \emph{adaptive sampling methods} for stochastic optimization problems. We distinguish throughout between the \emph{exact} population-level conditions used in the convergence analysis and the \emph{approximate} batch-based tests used in implementation. Although such methods have been extended to more general settings (see \Cref{sec:related} for details), here we focus on the unconstrained case.

\subsection{Norm Test}

\citet{byrd2012sample} proposed the \emph{norm test}, also called the \emph{norm condition}, based on the following observation.

\begin{proposition}
Let $x\in\RR^d$ satisfy $\nabla F(x)\neq 0$. If there exists $\eta\in[0,1)$ such that
\begin{equation}\label{eqn:condition}
    \delta_\calB(x) \coloneqq \|\nabla F_{\calB}(x) - \nabla F(x)\| \le \eta\|\nabla F(x)\|,
\end{equation}
then $-\nabla F_{\calB}(x)$ is a descent direction for $F$ at $x$.
\end{proposition}
\begin{proof}
A vector $d$ is a descent direction for $F$ at $x$ if and only if $\dotp{\nabla F(x)}{d}<0$. Thus it suffices to show that $\dotp{\nabla F_\calB(x)}{\nabla F(x)}>0$. If \eqref{eqn:condition} holds, then
\[
    \|\nabla F_{\calB}(x) - \nabla F(x)\|^2
    = \|\nabla F_{\calB}(x)\|^2 - 2\dotp{\nabla F_\calB(x)}{\nabla F(x)} + \|\nabla F(x)\|^2 
    \le \eta^2\|\nabla F(x)\|^2,
\]
which implies
\[
    2\dotp{\nabla F_\calB(x)}{\nabla F(x)}
    \ge (1-\eta^2)\|\nabla F(x)\|^2 + \|\nabla F_{\calB}(x)\|^2 > 0,
\]
because $\eta<1$ and $\nabla F(x)\neq 0$. Therefore $-\nabla F_{\calB}(x)$ is a descent direction for $F$ at $x$.
\end{proof}

For iteration $k\in\NN^*$, we define the filtration $\calF_k\coloneqq\sigma(\{x_1,\calB_1,\ldots,\calB_{k-1}\})$, which contains the randomness revealed before drawing the batch $\calB_k$ at iteration $k$. We abbreviate $\Ex_k[\cdot]\coloneqq\Ex[\cdot\mid \calF_k]$. The exact counterpart used in the convergence analysis is the \emph{exact variance norm test}
\begin{equation}\label{eqn:exact_norm}
    \Ex_k\left[\|\nabla F_{\calB_k}(x_k) - \nabla F(x_k)\|^2\right]
    \le \eta^2 \|\nabla F(x_k)\|^2.
\end{equation}
This is an expectation-based analogue of \eqref{eqn:condition}. It is not a pointwise guarantee.

When $\calB$ is sampled uniformly without replacement from $\setn$, the left-hand side of \eqref{eqn:exact_norm} can be estimated from the batch by
\begin{equation*}
    \hat\delta_\calB(x)^2\coloneqq\frac{n-b}{nb}\Var_{i\in\calB}(\nabla f(x; \xi_i)),
\end{equation*}
where, for any vector-valued function $h\colon\RR^d\times\calZ\to\RR^d$, the sample variance is defined by
\begin{equation}\label{eqn:def_var}
    \Var_{i\in\calB}(h(x; \xi_i))
    \coloneqq \frac{1}{b-1}\sum_{i\in\calB}\,\left\|h(x; \xi_i) - \frac1b\sum_{j\in\calB} h(x; \xi_j)\right\|^2.
\end{equation}
In the large-data regime, or under sampling with replacement, one typically drops the finite-population correction and uses the \emph{approximate norm test}
\begin{equation}\label{eqn:approx_norm}
    \frac{\Var_{i\in\calB_k}(\nabla f(x_k; \xi_i))}{b_k}
    \le \eta^2 \|\nabla F_{\calB_k}(x_k)\|^2.
\end{equation}
If \eqref{eqn:approx_norm} fails, a common monotone update is
\[
    b_{k+1}
    = \max\left\{b_k,\left\lceil \frac{\Var_{i\in\calB_k}(\nabla f(x_k; \xi_i))}{\eta^2 \|\nabla F_{\calB_k}(x_k)\|^2} \right\rceil\right\}.
\]
This is the smallest new batch size that would satisfy \eqref{eqn:approx_norm} if the estimated variance and batch-gradient norm were unchanged. In practice, one uses the statistics computed from the current batch to choose $b_{k+1}$ and then draws the next batch at that size, without rechecking the test on an enlarged batch at $x_k$.

\subsection{Inner Product Test}

\citet{bollapragada2018adaptive} observed that the norm test often increases the batch size too aggressively, and proposed the \emph{inner product test} as a milder alternative. In the conditionally i.i.d.~sampling model, the \emph{exact variance inner product test} requires that there exists $\vartheta>0$ such that
\begin{equation}\label{eqn:exact_inner_prod}
    \frac{1}{b_k}\Ex_k\left[\left(\dotp{\nabla f(x_k; \xi)}{\nabla F(x_k)} - \|\nabla F(x_k)\|^2\right)^2\right]
    \le \vartheta^2\|\nabla F(x_k)\|^4,
\end{equation}
where $\xi$ denotes a generic sample drawn according to the batch-sampling scheme at iteration $k$. Under conditional independence, this controls the variance of the projection $\dotp{\nabla F_{\calB_k}(x_k)}{\nabla F(x_k)}$.

Since $\nabla F(x_k)$ is not available, implementations usually replace it with $\nabla F_{\calB_k}(x_k)$ and use the \emph{approximate inner product test}
\begin{equation}\label{eqn:approx_inner_prod}
    \frac{\Var_{i\in\calB_k}(\dotp{\nabla f(x_k; \xi_i)}{\nabla F_{\calB_k}(x_k)})}{b_k}
    \le \vartheta^2\|\nabla F_{\calB_k}(x_k)\|^4,
\end{equation}
where the variance on the left-hand side is computed using \eqref{eqn:def_var}.

To rule out the possibility that $\nabla F_{\calB_k}(x_k)$ is nearly orthogonal to $\nabla F(x_k)$, \citet{bollapragada2018adaptive} also introduced the \emph{exact variance orthogonality test}: for $\nabla F(x_k)\neq 0$, there exists $\nu>0$ such that
\begin{equation}\label{eqn:exact_ortho}
    \frac{1}{b_k}\Ex_k\left[\left\|\nabla f(x_k;\xi) - \frac{\dotp{\nabla f(x_k; \xi)}{\nabla F(x_k)}}{\|\nabla F(x_k)\|^2}\nabla F(x_k)\right\|^2\right]
    \le \nu^2\|\nabla F(x_k)\|^2.
\end{equation}
The practical approximation, valid when $\nabla F_{\calB_k}(x_k)\neq 0$, is
\begin{equation}\label{eqn:approx_ortho}
    \frac{1}{b_k}\Var_{i\in\calB_k}\left( \nabla f(x_k; \xi_i) - \frac{\dotp{\nabla f(x_k; \xi_i)}{\nabla F_{\calB_k}(x_k)}}{\|\nabla F_{\calB_k}(x_k)\|^2} \nabla F_{\calB_k}(x_k)\right)
    \le \nu^2\|\nabla F_{\calB_k}(x_k)\|^2,
\end{equation}
where the variance on the left-hand side is again computed via \eqref{eqn:def_var}. The conditions \eqref{eqn:approx_inner_prod} and \eqref{eqn:approx_ortho} together form the \emph{approximate augmented inner product test}.

If either approximate condition fails, one may update the next batch size by
\begin{multline*}
    b_{k+1}
    = \max\Biggl\{b_k,\Biggl\lceil \max\Biggl\{ \frac{\Var_{i\in\calB_k}(\dotp{\nabla f(x_k; \xi_i)}{\nabla F_{\calB_k}(x_k)})}{\vartheta^2\|\nabla F_{\calB_k}(x_k)\|^4}, \\
    \frac{\Var_{i\in\calB_k}\left( \nabla f(x_k; \xi_i) - \frac{\dotp{\nabla f(x_k; \xi_i)}{\nabla F_{\calB_k}(x_k)}}{\|\nabla F_{\calB_k}(x_k)\|^2} \nabla F_{\calB_k}(x_k)\right)}{\nu^2\|\nabla F_{\calB_k}(x_k)\|^2} \Biggr\} \Biggr\rceil \Biggr\},
\end{multline*}
and then use a batch of that size at the next iteration. As with the norm test, this update is heuristic and is based on the current batch statistics. The constants $(\vartheta,\nu)\in\Rpp^2$ must be chosen in practice.

In large-scale implementations, one may also use the inner product test alone, since computing the quantities in \eqref{eqn:approx_ortho} introduces additional overhead and near-orthogonality has not been observed in practice in \citet{bollapragada2018adaptive}. In that case, the batch size is updated according to
\[
    b_{k+1}
    = \max\left\{b_k,\left\lceil \frac{\Var_{i\in\calB_k}(\dotp{\nabla f(x_k; \xi_i)}{\nabla F_{\calB_k}(x_k)})}{\vartheta^2\|\nabla F_{\calB_k}(x_k)\|^4}\right\rceil\right\}.
\]

\subsection{Adaptive Sampling Methods for Adaptive Gradient Methods}

We focus on two simple adaptive gradient methods, \AdaGrad \citep{duchi2011adagrad,mcmahan2010adaptive} and \AdaGradNorm \citep{streeter2010less}, whose step sizes are computed adaptively from the magnitudes of previous stochastic gradients.

\AdaGrad was proposed for online convex optimization and takes the form
\begin{equation}\label{eqn:AdaGrad}
    (\forall k\in\NN^*)\quad
    v_k = v_{k-1} + g_k^2, \quad
    x_{k+1} = x_k - \alpha g_k \odot v_k^{-\half},
\end{equation}
where $g_k\coloneqq\nabla F_{\calB_k}(x_k)$, $\alpha>0$ is a constant step size, $\odot$ denotes the Hadamard product, and the powers are taken coordinate-wise. Since $(v_k)_{k\in\NN}\subset\Rpp^d$, \AdaGrad uses adaptive coordinate-wise step sizes.

\AdaGradNorm, also known as \SGD with \AdaGrad step sizes, is the scalar variant of \AdaGrad:
\begin{equation}\label{eqn:AdaGrad-Norm}
    (\forall k\in\NN^*)\quad
    v_k = v_{k-1} + \|g_k\|^2, \quad
    x_{k+1} = x_k - \alpha g_k / \sqrt{v_k},
\end{equation}
where $(v_k)_{k\in\NN}\subset\Rpp$ is a sequence of positive scalars. The scalar step size $\alpha/\sqrt{v_k}$ makes \AdaGradNorm easier to analyze \citep{ward2019adagrad,ward2020adagrad,faw2022power,attia2023sgd}.

The norm test and the augmented inner product test were originally developed for \SGD, but they depend only on the discrepancy between full gradients and batch gradients and are therefore largely optimizer-agnostic. Thus \AdAdaGrad and \AdAdaGradNorm are obtained by combining the same batch-size rules with the updates \eqref{eqn:AdaGrad} and \eqref{eqn:AdaGrad-Norm}. The complete pseudocode is given in \Cref{alg:full}.

\begin{algorithm}[htbp]
\caption{Adaptive Batch Size Schemes for (Adaptive) Stochastic Gradient Methods (\AdaSGD, \AdAdaGrad, and \AdAdaGradNorm)}
\label{alg:full}
\begin{algorithmic}
\Require $x_1\in\RR^d$; $v_0\in\Rpp^d$ for \AdAdaGrad or $v_0\in\Rpp$ for \AdAdaGradNorm; $\calD=\{\xi_i\}_{i\in\setn}\subset\calZ$; either $\eta\in(0,1)$ (norm test) or $(\vartheta,\nu)\in\Rpp^2$ (augmented inner product test); total number of processed samples $N\in\NN^*$; step counter $k=1$; processed-samples counter $i=0$; initial batch size $2\le b_1\ll n$; step size sequence $(\alpha_k)_{k\in\NN^*}$ for \AdaSGD or step size $\alpha>0$ for \AdAdaGrad and \AdAdaGradNorm

\While{$i < N$}

    \State Sample a minibatch $\calB_k$ of size $b_k$ from $\calD$ according to the chosen sampling scheme
    \State Compute the batch gradient $g_k \coloneqq \nabla F_{\calB_k}(x_k) = \frac1{b_k}\sum_{i\in\calB_k}\nabla f(x_k; \xi_i)$

    \If{norm test}   
        \State Compute $\Var_{i\in\calB_k}(\nabla f(x_k; \xi_i))$
    
        \If{coordinate-wise}
            \State Compute the coordinate-wise norm test statistics $          \sfT_j(x_k; \calB_k, \eta) \coloneqq \frac{1}{b_k-1}\sum_{i\in\calB_k} \left(\partial_j f(x_k; \xi_i) - \partial_j F_{\calB_k}(x_k)\right)^2 \big/ \left(\eta^2 (\partial_j F_{\calB_k}(x_k))^2\right),
            \quad j\in\setd$        
        \State Compute the aggregate coordinate-wise norm test statistic $\sfT = \max_{j\in\setd} \sfT_j$
        \Else        
        \State Compute the norm test statistic 
        $\sfT\equiv\sfT(x_k; \calB_k, \eta) \coloneqq\Var_{i\in\calB_k}(\nabla f(x_k; \xi_i)) \big/\left(\eta^2 \|g_k\|^2 \right)$
        \EndIf        		
        \EndIf
        \If{augmented inner product test}
        \State Compute the variance of the inner product between the batch per-sample gradients and the batch gradient $\Var_{i\in\calB_k}(\dotp{\nabla f(x_k; \xi_i)}{\nabla F_{\calB_k}(x_k)})$
        \State Compute the inner product test statistic
        $\sfT_{\mathrm{ip}}(x_k; \calB_k, \vartheta) \coloneqq\Var_{i\in\calB_k}(\dotp{\nabla f(x_k; \xi_i)}{\nabla F_{\calB_k}(x_k)})\big/\left(\vartheta^2 \|\nabla F_{\calB_k}(x_k)\|^4 \right)$
        \State Compute the variance of the discrepancy of orthogonality between the batch per-sample gradients and the batch gradient 
        \[\sfV(x_k; \calB_k) \coloneqq\Var_{i\in\calB_k}\left( \nabla f(x_k; \xi_i) - \frac{\dotp{\nabla f(x_k; \xi_i)}{\nabla F_{\calB_k}(x_k)}}{\|\nabla F_{\calB_k}(x_k)\|^2} \nabla F_{\calB_k}(x_k)\right)\]
        \State Compute the orthogonality test statistic 
        \[\hspace*{25mm}\sfT_{\mathrm{ortho}}(x_k; \calB_k, \nu) \coloneqq\sfV(x_k; \calB_k)/\left(\nu^2 \|\nabla F_{\calB_k}(x_k)\|^2 \right)\]
        \State Compute the augmented inner product test statistic $\sfT=\max\{\sfT_{\mathrm{ip}}, \sfT_{\mathrm{ortho}}\}$
        \EndIf

        \State $b_{k+1} = \max\{\lceil\sfT\rceil, b_k\}$
        \State $x_{k+1} = x_k - \alpha_k\nabla F_{\calB_k}(x_k)$ \Comment{\AdaSGD}
        \State or
        \State $v_k = v_{k-1} + \|\nabla F_{\calB_k}(x_k)\|^2$ and $x_{k+1} = x_k - \alpha \nabla F_{\calB_k}(x_k)/ \sqrt{v_k}$ \Comment{\AdAdaGradNorm}
        \State or
        \State $v_k = v_{k-1} + \nabla F_{\calB_k}(x_k)^2$ and $x_{k+1} = x_k - \alpha \nabla F_{\calB_k}(x_k)\odot v_k^{-\half}$ \Comment{\AdAdaGrad}    
        \State $i \leftarrow i+b_k$
        \State $k \leftarrow k+1$                
        \EndWhile
    \end{algorithmic}
\end{algorithm}

Whenever a denominator in a test statistic vanishes, the corresponding condition is understood in the limiting sense that it can hold only if the numerator also vanishes.

The coordinate-wise nature of the adaptive step sizes in \AdaGrad motivates a coordinate-wise variant of \AdAdaGrad. This leads to the coordinate-wise exact variance norm test: for each $j\in\setd$,
\begin{equation}\label{eqn:co_exact_norm}
    \Ex_k\left[ \left(\partial_j F_{\calB_k}(x_k)-\partial_j F(x_k) \right)^2\right]
    \le \eta^2 (\partial_j F(x_k))^2,
\end{equation}
with approximate form
\[
    \frac{1}{b_k(b_k-1)}\sum_{i\in\calB_k} \left(\partial_j f(x_k; \xi_i) - \partial_j F_{\calB_k}(x_k)\right)^2
    \le \eta^2 (\partial_j F_{\calB_k}(x_k))^2.
\]
This coordinate-wise condition is stronger than the scalar norm test and is the one used later to analyze \AdAdaGrad. By contrast, the inner product and orthogonality tests intrinsically couple coordinates through inner products, so straightforward coordinate-wise analogues are not immediate.

\section{Convergence Analysis}
\label{sec:conv}

We provide two sets of convergence-rate results for \AdAdaGrad and \AdAdaGradNorm. Under $L$-Lipschitz smoothness, we establish convergence-rate results for \AdAdaGradNorm with either the norm test or the augmented inner product test, and for \AdAdaGrad with a coordinate-wise norm test. We then analyze \AdAdaGradNorm under generalized smoothness. We first record the smoothness assumptions used below.

\begin{assumption}[$L$-Lipschitz smoothness]\label{ass:grad_Lip}
    The function $F\colon\RR^d\to\RR$ is continuously differentiable, bounded below by $F^\star\coloneqq\inf_{x\in\RR^d} F(x)\in\RR$, and its gradient $\nabla F$ is $L$-Lipschitz continuous with constant $L>0$, that is, for any $x,y\in\RR^d$,
    \[
        \|\nabla F(x)-\nabla F(y)\| \le L\|x-y\|.
    \]
\end{assumption}

The uniform smoothness condition is often excessively restrictive in practice. When $F$ is twice continuously differentiable, \Cref{ass:grad_Lip} is equivalent to the uniform Hessian bound $\|\nabla^2 F(x)\|_{\mathrm{op}}\le L$ for all $x\in\RR^d$ (equivalently, $-LI_d\preccurlyeq\nabla^2 F(x)\preccurlyeq LI_d$). For example, \citet{zhang2020improved,zhang2020why} showed numerically that transformer architectures in language models exhibit loss landscapes that either do not satisfy the Lipschitz smoothness assumption or have very large Lipschitz constants $L$. To address this issue, \citet{zhang2020why} proposed the following weaker generalized smoothness condition:
\begin{assumption}[$(L_0,L_1)$-smoothness]\label{ass:gen_smooth}
    The function $F\colon\RR^d\to\RR$ is continuously differentiable, bounded below by $F^\star\coloneqq\inf_{x\in\RR^d} F(x)\in\RR$, and satisfies
    \[
        \|\nabla F(x)-\nabla F(y)\| \le (L_0 + L_1\|\nabla F(x)\|)\|x-y\|,
    \]
    for any $x,y\in\RR^d$, where $(L_0,L_1)\in\Rp^2$.
\end{assumption}

In addition to these smoothness assumptions, the adaptive sampling tests induce a control on the second moment of the stochastic gradient along the iterates. Using the nomenclature of \citet{khaled2023better}, we introduce the expected strong growth condition \citep{vaswani2019fast} below.
\begin{definition}[Expected strong growth]
    Let $\calB\subset\setn$ denote a random batch. The \emph{expected strong growth} (E-SG) condition is given by
    \[
        (\forall x\in\RR^d)\qquad \Ex[\|\nabla F_{\calB}(x)\|^2] \le \tau\|\nabla F(x)\|^2
    \]
    for some constant $\tau>0$, where the expectation is taken with respect to the batch-sampling randomness. In our analysis, it suffices to require this condition only along the iterates, that is,
    \begin{equation}\label{eqn:exp_strong_growth}
        (\forall k\in\NN^*)\quad\Ex_k[\|\nabla F_{\calB_k}(x_k)\|^2] \le \tau\|\nabla F(x_k)\|^2,
    \end{equation}
    for some constant $\tau>0$, where $(x_k)_{k\in\NN^*}$ are the iterates generated by a stochastic gradient method.
\end{definition}

\paragraph{Exact-test convention for the convergence analysis.}
The convergence results in this section are stated for idealized exact-test versions of the adaptive batch-size schemes. Specifically, at iteration $k$, the batch size is assumed to be chosen so that the corresponding population-level variance condition holds at the current iterate $x_k$ before the stochastic update is taken. This convention is used to isolate the convergence properties of the adaptive-sampling principle from the additional statistical error introduced by estimating the test quantities from a finite minibatch. The empirical algorithms in \Cref{sec:expt} use batch-based approximate tests, where the statistic computed from the current minibatch is used to choose the next batch size. Thus, the results below should be interpreted as convergence guarantees for the exact-test adaptive-sampling principle rather than for every implementation detail of the approximate empirical rule. 

\begin{proposition}[Informal]
    \label{prop:tests}
    Suppose that the batch gradient is conditionally unbiased at each iteration. If the exact variance norm test holds with constant $\eta\in(0,1)$, then the iteration-wise E-SG condition \eqref{eqn:exp_strong_growth} holds with $\tau=1+\eta^2$. Likewise, if the samples in $\calB_k$ are conditionally i.i.d.~and $\nabla F(x_k)\neq 0$, then the exact variance augmented inner product test with constants $(\vartheta,\nu)\in\Rpp^2$ implies \eqref{eqn:exp_strong_growth} with $\tau=1+\vartheta^2+\nu^2$.
\end{proposition}

A more precise statement and its proof are given in \Cref{sec:proofs}; see \Cref{prop:norm,prop:inner_prod}. Although the expected strong growth condition is often studied in overparameterized models \citep{vaswani2019fast} and is widely used in deep learning, the constant $\tau$ is usually problem-specific and unknown. In contrast, the adaptive sampling methods only require the corresponding bound to hold along the realized iterates $(x_k)_{k\in\NN^*}$, and the resulting value of $\tau$ is explicitly determined by the test parameters. When an approximate test fails, the batch size is increased according to the adaptive-sampling rule from \Cref{sec:adaptive_sampling}; under the approximation described there, this is intended to make the test hold for the enlarged batch, rather than to provide a separate guarantee at the next iterate.

\paragraph{Comparison with existing convergence rates.}
Before presenting our main results, we summarize in \Cref{tab:rate-comparison} how the proposed guarantees compare with representative convergence results for \SGD, adaptive sampling methods, \AdaGrad, and \AdaGradNorm on smooth nonconvex objectives. The comparison is qualitative: different results use different stochastic-gradient assumptions, and the rates below should be interpreted together with the corresponding noise or sampling conditions. In particular, the $\scrO(1/K)$ rates obtained in this paper rely on exact adaptive sampling tests, which imply an expected strong growth condition along the iterates. They should not be interpreted as unconditional improvements over the standard $\scrO(1/\sqrt{K})$ rates under bounded variance.

\begin{table}[h]
\centering
\caption{Comparison of convergence guarantees for stochastic gradient methods on smooth nonconvex objectives. Here, $K$ denotes the number of iterations and $\delta\in(0,1)$ denotes the failure-probability parameter in the high-probability guarantees; equivalently, the corresponding bounds hold with probability at least $1-\delta$. The notation $\widetilde{\scrO}$ hides logarithmic factors and problem-dependent constants.}
\label{tab:rate-comparison}
\scriptsize
\begin{tabular}{llll}
    \toprule
    Method & Sampling / noise condition & Guarantee type & Rate \\
    \midrule
    \SGD & bounded variance & expectation & $\mathcal{O}(K^{-1/2})$ \\[1mm]
    \SGD with adaptive sampling & exact adaptive sampling test & expectation & $\scrO(K^{-1})$ \\[1mm]
    \multirow{2}{*}{\AdaGrad} & \multirow{2}{*}{standard stochastic gradient assumptions} & \multirow{2}{*}{\shortstack{expectation or \\high probability}} & \multirow{2}{*}{$\widetilde{\scrO}(K^{-1/2})$} \\[1mm]
    &  &  &  \\
    \multirow{2}{*}{\AdaGradNorm} & \multirow{2}{*}{standard stochastic-gradient assumptions} & \multirow{2}{*}{\shortstack{expectation or \\high probability}} & \multirow{2}{*}{$\widetilde{\scrO}(K^{-1/2})$} \\[1mm]
    &  &  &  \\
    \AdAdaGradNorm & exact norm or augmented inner-product test & high probability & $\scrO(K^{-1}\delta^{-2})$ \\[1mm]
    \AdAdaGrad & coordinate-wise exact norm test & high probability & $\scrO(K^{-1}\delta^{-2})$ \\[1mm]
    \multirow{2}{*}{\shortstack{\AdAdaGradNorm under \\$(L_0,L_1)$-smoothness}} & \multirow{2}{*}{exact norm or augmented inner-product test} & \multirow{2}{*}{high probability} & \multirow{2}{*}{$\scrO(K^{-1}\delta^{-2})$} \\ 
    &  &  &  \\
    \bottomrule
\end{tabular}
\end{table}

\subsection{Convergence Results}
We first establish high-probability convergence rates for the adaptive batch size schemes used with \AdaGradNorm and \AdaGrad, substantially extending existing convergence-rate results in expectation for \SGD; see, e.g., \citet{de2016big,de2017automated,bollapragada2018adaptive}. 

\paragraph{Technical challenges beyond \SGD.}
The main difficulty relative to adaptive sampling for \SGD is that the stochastic gradient appears both in the update direction and in the adaptive denominator. For \AdaGradNorm, the update uses $g_k/\sqrt{v_k}$ with $v_k=v_{k-1}+\|g_k\|^2$, so the random numerator $g_k$ and the random denominator $\sqrt{v_k}$ are coupled. Consequently, conditional unbiasedness of $g_k$ does not directly imply an unbiased descent direction after adaptive normalization. Our analysis handles this coupling by decomposing
\[
\left\langle \nabla F(x_k),\mathbb{E}_k\left[\frac{g_k}{\sqrt{v_k}}\right]\right\rangle
=
\frac{\|\nabla F(x_k)\|^2}{\sqrt{v_{k-1}}}
+
\left\langle \nabla F(x_k),\mathbb{E}_k\left[\left(\frac{1}{\sqrt{v_k}}-\frac{1}{\sqrt{v_{k-1}}}\right)g_k\right]\right\rangle,
\]
and bounding the second term through a potential-difference argument. The remaining second-order terms are controlled by logarithmic \AdaGrad-type sums such as $\sum_{k=1}^K\|g_k\|^2/v_k\le \log(v_K)-\log(v_0)$. For coordinate-wise \AdaGrad, the same issue occurs separately in each coordinate, which motivates the coordinate-wise norm test used in the convergence theorem.

\AdaGradNorm combined with either the norm test or the augmented inner product test, with any constant initial step size $\alpha>0$, enjoys the following high-probability convergence bound for nonconvex functions. Before stating the convergence result for \AdAdaGradNorm, we emphasize that the theorem applies to the exact variance tests : the exact variance norm test or the augmented inner product test is assumed to hold at the iterate $x_k$ used for the update. The batch-based implementation used in the experiments is an approximate version of this idealized rule.

\begin{theorem}[\AdAdaGradNorm]\label{thm:AdaGrad-Norm}
    Suppose that \Cref{ass:grad_Lip} holds. Let $(x_k)_{k\in\NN^*}$ be the \AdaGradNorm iterates \eqref{eqn:AdaGrad-Norm} with any step size $\alpha>0$, where the batch sizes $(b_k)_{k\in\NN^*}$ are chosen so that either the (exact variance) norm test \eqref{eqn:exact_norm} with constant $\eta\in(0,1)$ or the (exact variance) augmented inner product test \eqref{eqn:exact_inner_prod} and \eqref{eqn:exact_ortho} with constants $(\vartheta,\nu)\in\Rpp^2$ is satisfied at each iteration. Fix $K\in\NN^*$, $\delta\in(0,1)$, and $\rho\in(1,\infty)$, and set
    \[
        \tau \coloneqq
        \begin{cases}
            1+\eta^2, & \text{for the norm test},\\
            1+\vartheta^2+\nu^2, & \text{for the augmented inner product test}.
        \end{cases}
    \]
    Define
    \[
        c_1 \coloneqq \frac{2}{\alpha(1-\rho^{-1})}\left(F(x_1)-F^\star + \frac{\tau\alpha\|\nabla F(x_1)\|^2}{2\sqrt{v_0}} + \frac{\tau L^2\alpha^3(1+\rho\tau)}{\sqrt{v_0}} - \frac{L\alpha^2}{2}\log(v_0)\right),
    \]
    \[
        c_1^+ \coloneqq \max\{c_1,0\},
        \qquad
        c_2 \coloneqq \frac{L\alpha}{1-\rho^{-1}},
        \qquad
        c_3 \coloneqq 2\sqrt{v_0} + 2\tau c_1^+ + 8\tau c_2\log(\tau c_2+1).
    \]
    Then, with probability at least $1-\delta$, we have
    \[
        \min_{k\in\setK}\norm{\nabla F(x_k)}^2 \le \frac{c_3(c_1^+ + 2c_2\log c_3)}{K\delta^2}.
    \]
\end{theorem}

The proof of \Cref{thm:AdaGrad-Norm} is technical, and its full details are given in \Cref{subsubsec:proof_thm6}. 

\paragraph{Choice of $\alpha$ and $\rho$ in \Cref{thm:AdaGrad-Norm}.}
\Cref{thm:AdaGrad-Norm} holds for any constant $\alpha>0$ under $L$-Lipschitz smoothness; no upper bound on $\alpha$ is required for the validity of the statement. However, the constants in the bound depend on $\alpha$, and an excessively large $\alpha$ may make the bound loose or practically uninformative. The parameter $\rho>1$ is not an algorithmic parameter. It is introduced only in the proof through Young-type inequalities and controls a trade-off among the constants. For a simple interpretation one may fix, for example, $\rho=2$; optimizing $\rho$ can slightly improve the displayed constant but does not affect the asymptotic rate.

To prove the convergence of \AdaGrad, which has coordinate-wise adaptive step sizes, we need the E-SG condition to hold coordinate-wise along the iterates as well. This indeed holds when we invoke a coordinate-wise version of the norm test, as stated in the following proposition.
\begin{proposition}[Coordinate-wise expected strong growth]\label{prop:co_E-SG}
    Suppose that, for every iteration $k\in\NN^*$, the batch gradient $\nabla F_{\calB_k}(x_k)$ is conditionally unbiased, that is,
    \[
        \Ex_k[\nabla F_{\calB_k}(x_k)] = \nabla F(x_k),
    \]
    and the coordinate-wise exact variance norm test \eqref{eqn:co_exact_norm} holds with constant $\eta\in(0,1)$. Then the \emph{coordinate-wise E-SG} condition holds at each iteration, that is, for all $(k,j)\in\NN^*\times\setd$,
    \begin{equation}\label{eqn:co_exp_strong_growth}
        \Ex_k\left[\left(\partial_j F_{\calB_k}(x_k)\right)^2\right]
        \le (1+\eta^2)(\partial_j F(x_k))^2.
    \end{equation}
\end{proposition}

\begin{proof}
    Fix $(k,j)\in\NN^*\times\setd$. Since
    \[
        \Ex_k[\partial_j F_{\calB_k}(x_k)] = \partial_j F(x_k),
    \]
    we have
    \[
        \Ex_k\left[\left(\partial_j F_{\calB_k}(x_k)-\partial_j F(x_k)\right)^2\right]
        = \Ex_k\left[\left(\partial_j F_{\calB_k}(x_k)\right)^2\right] - (\partial_j F(x_k))^2.
    \]
    Combining this identity with \eqref{eqn:co_exact_norm} yields \eqref{eqn:co_exp_strong_growth}.
\end{proof}

Then \AdaGrad with the coordinate-wise norm test enjoys a similar high-probability convergence guarantee. For \AdAdaGrad, the following result is proved under the coordinate-wise exact variance norm test. This condition is stronger than the standard norm test used in our neural network experiments, but it is aligned with the coordinate-wise adaptive preconditioner in \AdaGrad and yields the coordinate-wise expected strong-growth control needed in the proof. 

\begin{theorem}[\AdAdaGrad]\label{thm:AdaGrad}
    Suppose that \Cref{ass:grad_Lip} holds. Let $(x_k)_{k\in\NN^*}$ be the \AdaGrad iterates \eqref{eqn:AdaGrad} with step size $\alpha>0$, where the batch gradients $g_k\coloneqq\nabla F_{\calB_k}(x_k)$ are conditionally unbiased,
    \[
        \Ex_k[g_k]=\nabla F(x_k), \qquad k\in\NN^*,
    \]
    and the batch sizes $(b_k)_{k\in\NN^*}$ are chosen so that the coordinate-wise exact variance norm test \eqref{eqn:co_exact_norm} holds with some $\eta\in(0,1)$ at each iteration. Then there exists a constant $C>0$, independent of $K\in\NN^*$ and $\delta\in(0,1)$, such that
    \[
        \Prob\left(\min_{k\in\setK}\norm{\nabla F(x_k)}^2 \le \frac{C}{K\delta^2}\right)\ge 1-\delta.
    \]
\end{theorem}
   
The above theorem establishes a sublinear convergence rate (with high probability) for \AdAdaGrad on nonconvex functions, whereas such a rate in expectation was established for \SGD in \citet{bollapragada2018adaptive}, Theorem 3.4. 

\paragraph{Scope of the \AdAdaGrad guarantee.}
Theorem~\ref{thm:AdaGrad} applies to \AdaGrad combined with the coordinate-wise exact variance norm test. This condition is natural for the coordinate-wise preconditioner in \AdaGrad and yields a coordinate-wise expected strong-growth bound. In numerical experiments (\Cref{sec:expt}), however, evaluating the coordinate-wise test is computationally expensive, and we therefore use the standard norm test as a practical approximation. Consequently, the \AdAdaGrad experiments should be viewed as empirical evidence for the practical standard-norm-test variant, rather than as a direct numerical verification of Theorem~\ref{thm:AdaGrad}.

Finally, after relaxing the uniform smoothness assumption, \AdAdaGradNorm still converges under $(L_0,L_1)$-smoothness. When $L_1=0$, this reduces to the Lipschitz-smooth case covered by \Cref{thm:AdaGrad-Norm}; below we therefore assume $L_1>0$. In this regime, the scale parameter $\alpha$ must be upper bounded, and hence one needs knowledge of $L_1$. The same exact-test convention is used in the generalized-smoothness setting below. 

\begin{theorem}[$(L_0,L_1)$-smooth \AdAdaGradNorm]
\label{thm:AdaGrad-Norm_gen_smooth}
    Suppose that \Cref{ass:gen_smooth} holds with $L_1>0$. Let $(x_k,v_k)_{k\in\NN^*}$ be generated by \eqref{eqn:AdaGrad-Norm} from some $v_0\in\Rpp$, fix $K\in\NN^*$ and $\delta\in(0,1)$, and define
    \[
        g_k \coloneqq \nabla F_{\calB_k}(x_k), \qquad k\in\setK.
    \]

    \begin{enumerate}[label=(\roman*)]
        \item for every $k\in\setK$, $g_k$ is conditionally unbiased and the exact variance norm test \eqref{eqn:exact_norm} holds with some $\eta\in(0,1)$, in which case $\tau\coloneqq 1+\eta^2$; or    
        \item for every $k\in\setK$, the samples in $\calB_k$ are conditionally i.i.d., $\nabla F(x_k)\neq 0$, and the exact variance inner product test \eqref{eqn:exact_inner_prod} and exact variance orthogonality test \eqref{eqn:exact_ortho} hold with some $(\vartheta,\nu)\in\Rpp^2$, in which case $\tau\coloneqq 1+\vartheta^2+\nu^2$.
    \end{enumerate}
    Let $(\rho_1,\rho_2,\omega)\in\Rpp^3$ satisfy
    \[
        \frac{1}{\rho_1}+\frac{\rho_1}{\rho_2}+2\omega<1,
    \]
    and suppose that
    \[
        \alpha\le \frac{1}{L_1}\min\left\{\frac{\omega}{2\rho_1\tau}, \sqrt{\frac{\omega}{2\rho_1\tau}}\right\}.
    \]
    Then there exists a constant $C>0$, depending only on the problem and algorithm parameters but independent of $K$ and $\delta$, such that
    \[
        \Prob\left(\min_{k\in\setK}\|\nabla F(x_k)\|^2 \le \frac{C}{K\delta^2}\right)\ge 1-\delta.
    \]
\end{theorem}

If $\nabla F(x_k)=0$ for some $k\in\setK$, then the conclusion is immediate. The same proof strategy suggests a corresponding generalized-smoothness result for \AdaGrad, but we do not state it here.

\subsection{Outlines of Proofs}
We now provide outlines of the proofs of the above theorems. Full proofs can be found in \Cref{sec:proofs}.

\begin{proof}[\Cref{thm:AdaGrad-Norm}]
    The proof first invokes the iteration-wise E-SG bound furnished by the exact test:
    \[
        \Ex_k[\|g_k\|^2] \le \tau\|\nabla F(x_k)\|^2,
    \]
    where $\tau=1+\eta^2$ for the norm test and $\tau=1+\vartheta^2+\nu^2$ for the augmented inner product test. Using the $L$-Lipschitz smoothness of $F$, we obtain
    \[
        \Ex_k[F(\xkpo)] \le F(x_k) - \alpha\lrdotp{\gradFxk}{\Ex_k\left[\frac{g_k}{\sqrt{v_k}}\right]} + \frac{L\alpha^2}{2}\Ex_k\left[\left\|\frac{g_k}{\sqrt{v_k}}\right\|^2\right].
    \]
    The inner product is decomposed as
    \[
        \lrdotp{\gradFxk}{\Ex_k\left[\frac{g_k}{\sqrt{v_k}}\right]}
        = \frac{\|\gradFxk\|^2}{\sqrt{\vkmo}}
        + \lrdotp{\gradFxk}{\Ex_k\left[\left(\frac{1}{\sqrt{v_k}}-\frac{1}{\sqrt{\vkmo}}\right)g_k\right]}.
    \]
    The second term is then controlled by a one-step potential difference $\varphi_{k-1}-\varphi_k$, where $\varphi_k=\|\nabla F(x_k)\|^2/\sqrt{v_k}$, together with a summable remainder of order $\|g_{k-1}\|^2/v_{k-1}^{3/2}$. After summing over $k$, the second-order term is handled through the logarithmic bound $\sum_{k=1}^K \|g_k\|^2/v_k \le \log v_K-\log v_0$, which yields, via Jensen's inequality, a uniform bound on $\Ex[\sqrt{v_K}]$. This in turn bounds $\sum_{k=1}^K \Ex[\|\nabla F(x_k)\|^2/\sqrt{v_{k-1}}]$, and a final Cauchy--Schwarz and Markov argument gives the stated high-probability estimate for $\min_{k\in\setK}\|\nabla F(x_k)\|^2$.
\end{proof}

\begin{proof}[\Cref{thm:AdaGrad}]
    The proof follows the same scheme, but coordinate-wise. Using the coordinate-wise exact variance norm test together with the conditional unbiasedness of $g_k$, one obtains
    \[
        \Ex_k[g_{k,j}^2] \le (1+\eta^2)(\partial_j F(x_k))^2,
        \qquad j\in\setd.
    \]
    The descent inequality becomes
    \[
        \Ex_k[F(\xkpo)]
        \le F(x_k)
        - \alpha\sum_{j=1}^d \frac{(\partial_jF(x_k))^2}{\sqrt{\vkmoj}}
        + \alpha\sum_{j=1}^d T_{k,j}
        + \frac{L\alpha^2}{2}\Ex_k\left[\left\|\frac{1}{\sqrt{v_k}}\odot g_k\right\|^2\right],
    \]
    where
    \[
        T_{k,j}\coloneqq
        \partial_jF(x_k)\Ex_k\left[\left(\frac{1}{\sqrt{\vkmoj}}-\frac{1}{\sqrt{\vkj}}\right)\gkj\right].
    \]
    Each $T_{k,j}$ is bounded by a coordinate-wise potential difference plus a summable remainder. Summing over $j$ and then over $k$ yields control of
    \[
        \sum_{k=1}^K \sum_{j=1}^d \Ex\left[\frac{(\partial_jF(x_k))^2}{\sqrt{v_{k-1,j}}}\right].
    \]
    The accumulated second-order terms are handled through the logarithmic bounds $\sum_{k=1}^K g_{k,j}^2/v_{k,j} \le \log v_{K,j}-\log v_{0,j}$, which then control $\sum_{j=1}^d \Ex[\log v_{K,j}]$ and hence $\Ex[\sum_{j=1}^d \sqrt{v_{K,j}}]$. The high-probability bound follows from the same final Cauchy--Schwarz and Markov argument as in the proof of \Cref{thm:AdaGrad-Norm}.
\end{proof}

\begin{proof}[\Cref{thm:AdaGrad-Norm_gen_smooth}]
    The proof parallels that of \Cref{thm:AdaGrad-Norm}, but begins with the generalized descent lemma. The step-size restriction ensures $L_1\|x_{k+1}-x_k\|\le 1$, so
    \[
        \Ex_k[F(\xkpo)] \le F(x_k) - \alpha\lrdotp{\gradFxk}{\Ex_k\left[\frac{g_k}{\sqrt{v_k}}\right]} + \frac{\alpha^2}{2}\left(L_0+L_1\|\gradFxk\|\right)\Ex_k\left[\left\|\frac{g_k}{\sqrt{v_k}}\right\|^2\right].
    \]
    As before, we split the inner product into a main descent term and an error term. The error term and the additional $L_1$-dependent second-order contribution are then combined and bounded by using the E-SG estimate, the generalized smoothness inequality, and the parameter restrictions involving $(\rho_1,\rho_2,\omega)$. This yields a descent inequality of the same general form as in the Lipschitz-smooth case, up to modified constants. Summing over $k$, controlling $\Ex[\sqrt{v_K}]$ through $\sum_{k=1}^K \|g_k\|^2/v_k \le \log v_K-\log v_0$ and Jensen's inequality, and finally applying Cauchy--Schwarz and Markov's inequality, we obtain the claimed bound $\min_{k\in\setK}\|\nabla F(x_k)\|^2=\scrO(1/(K\delta^2))$ with high probability.
\end{proof}

\section{Numerical Experiments}
\label{sec:expt}

We evaluate the performance of the norm test and the (augmented) inner product test with \textsc{AdaGrad}(-\textsc{Norm}) and \SGD for image classification, employing logistic regression (\Cref{subsec:log_reg}) and a three-layer CNN on the MNIST dataset \citep{lecun1998mnist}, as well as a three-layer CNN and \ResNet-18 \citep{he2016deep} on the CIFAR-10 dataset \citep{krizhevsky0209learning}. We note that training larger models often requires multiple workers and data parallelism, such as Distributed Data Parallel \citep[DDP;][]{li2020pytorch} and Fully Sharded Data Parallel \citep[FSDP;][]{zhao2023pytorch}. Extending and implementing our proposed schemes under data parallelism presents additional complexities and remains an area for future research \citep[see e.g.,][for recent results]{lau2024adaptive_lm}. We thus concentrate on smaller models and datasets with the goal of demonstrating the concept rather than achieving state-of-the-art results. Given the typically large number of parameters $d$, conducting coordinate-wise norm tests for \AdAdaGrad is not computationally practical, so the standard norm test is applied. For the purpose of numerical comparison, we also empirically evaluate the norm test with \Adam \citep{kingma2015} for training \ResNet-18 on the CIFAR-10 dataset, which is coined \AdAdam. 

For the implementation of the proposed schemes, per-sample gradients are computed using JAX-like composable function transforms \citep{jax2018github} called \href{https://pytorch.org/docs/master/func.html}{\texttt{torch.func}} in PyTorch 2.0+. Numerical experiments are carried out on workstations with NVIDIA RTX 2080Ti 11GB (for MNIST), A100 80GB (for CNN on CIFAR-10) and L40S 48GB (for \ResNet-18 on CIFAR-10) GPUs, with PyTorch 2.2.1 \citep{paszke2019pytorch} and Lightning Fabric 2.2.0 \citep{falcon2019pytorch}. The \AdaGradNorm implementation is taken from \citet{ward2019adagrad}. 
The code for the full implementation of the proposed schemes and reproducing the following training results is available at \url{https://github.com/timlautk/AdAdaGrad}. 
Further details of the experiments, such as the hyperparameter setting, and additional results can be found in \Cref{sec:expt_details}. 

\subsection{Multi-class Logistic Regression on MNIST}   
\label{subsec:log_reg}

We first apply our methods to a ten-class logistic-regression problem on the MNIST dataset, which has a smooth convex objective. Our experiments are conducted with an equal training budget of 6 million samples (equivalent to 100 epochs), setting a maximum batch size of 60,000 (i.e., the full batch) for all approaches. To highlight the adaptivity and flexibility of our proposed methods, we refrain from conducting an exhaustive search for optimal values of $\alpha$, $\eta$, and $\vartheta$, and we do not use learning-rate schedules across methods. The outcomes, including the number of iterations required (steps), average batch sizes (batch size), final training loss (loss), and final validation accuracy (accuracy), are reported in \Cref{table:log_reg}.

    \begin{table}[h!]
        \centering
        \caption{\small Multi-class logistic regression on MNIST}    
        \label{table:log_reg}
        \begin{tabular}{ccrrrr}
            \toprule
            Scheme & test & steps & batch size & loss & accuracy \\
            \midrule
            \AdaSGD & $\eta=0.10$ & 351 & 17131 &  1.04 &  0.82  \\
            \AdaSGD & $\eta=0.25$ & 1029  & 5831 &  0.67 &  0.86 \\
            \AdAdaGradNorm & $\eta=0.10$ & 596 & 10060 & 1.50  &  0.78  \\
            \AdAdaGradNorm & $\eta=0.25$  & 2462 & 2437 & 1.02 &  0.83  \\
            \AdAdaGrad & $\eta=0.10$ & \textbf{126} & \textbf{47282} & \textbf{0.54} & \textbf{0.88} \\
            \AdAdaGrad & $\eta=0.25$ & \textbf{274} & \textbf{21918} & \textbf{0.46} & \textbf{0.90} \\
            \AdaSGD & $\vartheta=0.01$ & 717 & 8362 & 0.76 & 0.85   \\
            \AdaSGD & $\vartheta=0.05$ & 1804 & 3327 & 0.54 & 0.88   \\
            \AdAdaGradNorm & $\vartheta=0.01$ & 1127 & 5322 & 1.28 &   0.80 \\
            \AdAdaGradNorm & $\vartheta=0.05$ & 5349 & 1122 & 0.80  &  0.85  \\
            \bottomrule
        \end{tabular}    
    \end{table} 	
    
\begin{figure}[h!]
        \centering
        \includegraphics[width=\textwidth]{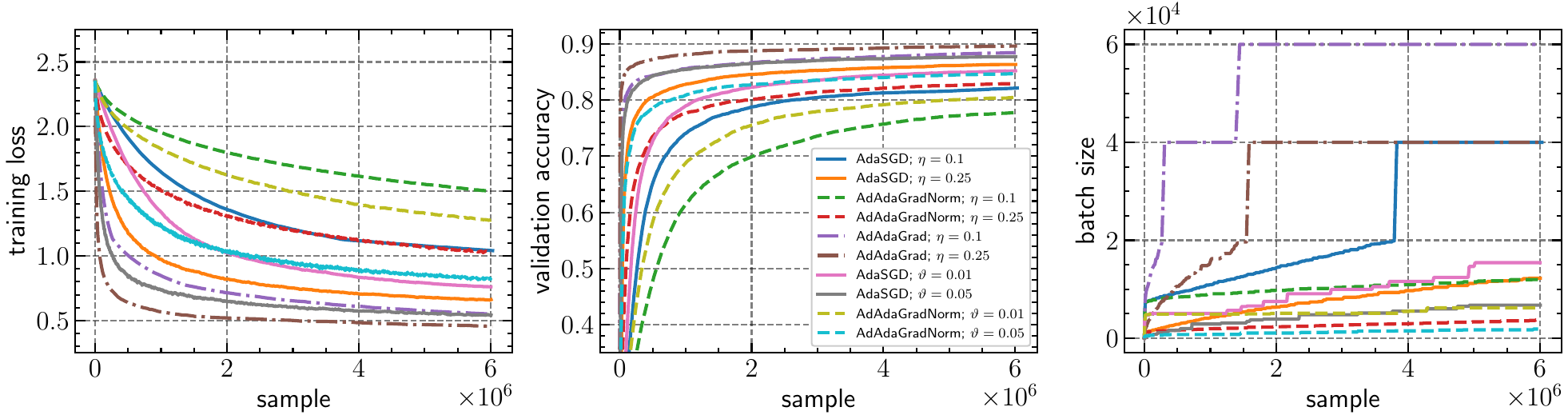}
    \caption{Training loss, validation accuracy, and batch-size curves (vs.~number of training samples) of \AdaSGD, \AdAdaGrad, and \AdAdaGradNorm for logistic regression on the MNIST dataset.}
        \label{fig:log_reg_mnist}
    \end{figure}

The results, together with the graphical representations in \Cref{fig:log_reg_mnist}, show that \AdAdaGrad, using the norm test with $\eta=0.25$, outperforms the other methods in terms of both training loss and validation accuracy. These findings underscore the importance of adaptive, and coordinate-wise, learning rates within adaptive sampling methods.
Despite the best overall performance of \AdAdaGrad with the norm test and $\eta=0.25$, the variant with $\eta=0.10$ is more hardware-efficient, with an average batch size exceeding 47,000 and requiring only 126 steps (i.e., gradient evaluations) to process all 6 million samples. This indicates an intrinsic trade-off between computational efficiency and model performance as measured by generalization. Interestingly, \AdAdaGradNorm, although theoretically simpler to analyze, may underperform relative to \AdaSGD (i.e., \SGD with adaptive batch size schemes) for this specific convex problem and hyperparameter setting. It is also worth noting that the norm test tends to increase batch sizes more aggressively than the inner product test, leading to more efficient use of available GPU memory.
We also observe from \Cref{fig:log_reg_mnist} that, under the norm test, the batch sizes of \AdAdaGrad grow much faster than those of \AdaSGD. We leave a systematic study and theoretical explanation of this phenomenon for future work.

\subsection{Three-layer CNN on MNIST}

\begin{figure}[t]
    \centering
    \includegraphics[width=\textwidth]{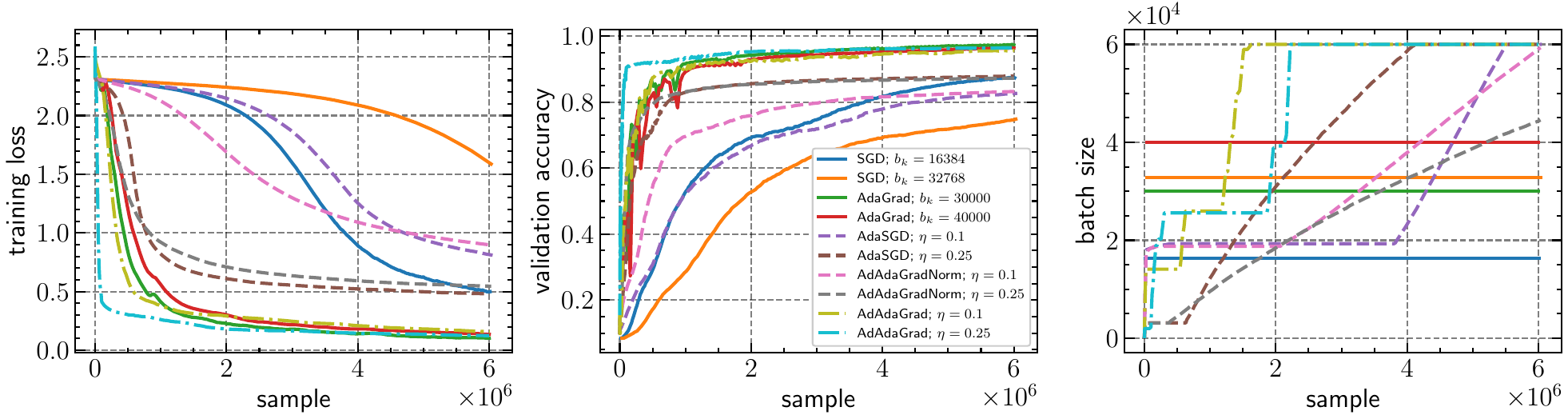}
    \caption{Training loss, validation accuracy, and batch sizes of \AdaSGD, \AdAdaGrad, and \AdAdaGradNorm for a three-layer CNN on the MNIST dataset.}
    \label{fig:cnn_mnist_adagrad}
\end{figure}

We next turn to the application of the proposed methods to training deep neural networks.
Specifically, we train a three-layer CNN on the MNIST classification problem, which has a nonconvex objective. Our experiments are conducted with an equal training budget of 6 million samples (equivalent to 100 epochs), setting a maximum batch size of 60,000 (i.e., the full batch) for all approaches. We measure training efficiency by the number of gradient steps rather than wall-clock time, which is device dependent \citep{shallue2019measuring}. Adaptive batch size methods begin with a small batch size of 8 and gradually increase to the maximum possible batch size of 60,000 (full batch). In \Cref{fig:cnn_mnist_adagrad}, \AdAdaGrad outperforms \AdAdaGradNorm and \AdaSGD in validation accuracy (generalization) by a clear margin. \AdAdaGradNorm achieves performance similar to that of \AdaSGD, despite having slightly higher training loss. We also observe from \Cref{table:cnn_mnist_full} that \AdAdaGrad using the norm test with $\eta=0.1$ achieves a validation accuracy of 96\% with only 149 iterations and an average batch size of more than 40,000. It uses full batches for the last 70\% of its training budget, taking full advantage of the available GPU memory. Referring to \Cref{table:cnn_mnist_full}, we observe the generalization gap between small constant batch sizes and large batch sizes, whereas using small batch sizes requires substantially more training time and a larger number of steps, leading to lower training efficiency. Our proposed methods are capable of balancing this trade-off between training efficiency and generalization by introducing adaptive batch size schemes, without the need for extensive tuning of learning rates or pre-specified learning-rate schedules.

We also compare the use of the norm test for \SGD and \AdaGrad. For this nonconvex problem, adaptive learning rates become crucial, as \AdaGrad converges much faster than \SGD with constant batch sizes. Using the norm test, we observe from \Cref{fig:cnn_mnist_adagrad} that \AdAdaGrad increases batch sizes more aggressively than \AdaSGD, indicating that shorter training time is required (see also \Cref{table:cnn_mnist_full}). This suggests the usefulness of adaptive batch size schemes based on the norm test for training nonconvex deep neural networks. 

\paragraph{Choice of constant-batch-size baselines.}
We include constant-batch-size baselines to compare the adaptive-batch methods against nonadaptive choices in comparable compute regimes. Since the proposed methods are designed to increase the batch size when the empirical variance test indicates that larger batches are beneficial, the most informative nonadaptive baselines are constant batch sizes in the large-batch regime reached by the adaptive methods. We therefore focus on constant batch sizes that span the practical range of batch sizes observed in the adaptive runs, rather than very small historical defaults such as $128$ or $256$. This comparison tests whether the adaptive rule can automatically select an effective large-batch regime without fixing the batch size in advance. 

\begin{table}[h!]
    \centering
    \caption{Three-layer CNN on MNIST}
    \label{table:cnn_mnist_full}
    \begin{tabular}{ccrrrrr}
        \toprule
        Scheme & test & steps & time (h) & batch size & loss & accuracy \\
        \midrule
        \SGD & N/A & 2929 & 6.77 & 2048 & 0.12 & 0.96 \\
        \SGD & N/A & 1464 & 3.38 & 4096 & 0.20 & 0.94 \\
        \SGD & N/A & 732 & 1.72 & 8192 & 0.32 & 0.91 \\
        \SGD & N/A & 366 & 0.90 & 16384 & 0.51 & 0.87 \\
        \SGD & N/A & 183 & 0.45 & 32768 & 1.54 & 0.75 \\
        \SGD & N/A & 99 & 0.27 & 60000 & 2.15 & 0.66 \\
        \AdaGrad & N/A & 2929 & 7.12 & 2048 & 0.02 & 0.99 \\
        \AdaGrad & N/A & 1464 & 3.60 & 4096 & 0.02 & 0.99 \\
        \AdaGrad & N/A & 732 & 1.82 & 8192 & 0.05 & 0.98 \\
        \AdaGrad & N/A & 366 & 0.92 & 16384 & 0.07 & 0.98 \\
        \AdaGrad & N/A & 199 & 0.52 & 30000 & 0.10 & 0.97 \\
        \AdaGrad & N/A & 183 & 0.47 & 32768 & 0.11 & 0.97 \\
        \AdaGrad & N/A & 149 & 0.36 & 40000 & 0.13 & 0.96 \\
        \AdaGrad & N/A & 99 & 0.29 & 60000 & 0.17 & 0.95 \\
        \AdaSGD & $\eta=0.10$ & 256 & 0.73 & 23546 &  0.79 & 0.83 \\
        \AdaSGD & $\eta=0.25$ & 383 & 1.05 & 15627 &  0.48 & 0.88 \\
        \AdAdaGradNorm & $\eta=0.10$ & 226 & 0.65 & 26567 &  0.88 &  0.83 \\
        \AdAdaGradNorm & $\eta=0.25$ & 435 & 1.27 & 13830 &  0.54 &  0.87 \\
        \AdAdaGrad & $\eta=0.10$ & \textbf{149} & \textbf{0.45} & \textbf{40057} & \textbf{0.15} & \textbf{0.96} \\
        \AdAdaGrad & $\eta=0.25$ & \textbf{198} & \textbf{0.58} & \textbf{30152} & \textbf{0.13} & \textbf{0.97} \\
        \AdAdaGrad & $\eta=0.50$ & \textbf{215} & \textbf{0.62} & \textbf{27940} & \textbf{0.11} & \textbf{0.97} \\
        \AdAdaGrad & $\eta=0.75$ & \textbf{271} & \textbf{0.79} & \textbf{22228} & \textbf{0.10} & \textbf{0.97} \\
        \AdaSGD & $\vartheta=0.01$ & 230 & 0.63 & 26078 & 0.98 & 0.80 \\
        \AdaSGD & $\vartheta=0.05$ & 411 & 1.17 & 14593 & 0.45 & 0.88 \\
        \AdAdaGradNorm & $\vartheta=0.01$ & 241 & 0.70 & 24872 & 0.83 & 0.84 \\
        \AdAdaGradNorm & $\vartheta=0.05$ & 528 & 1.44 & 11365 & 0.50 & 0.88 \\
        \bottomrule
    \end{tabular}
\end{table}

\subsection{Three-layer Convolutional Neural Network on CIFAR-10}
\label{subsec:cnn_cifar10}

\begin{figure}[h!]
        \centering
        \includegraphics[width=0.8\textwidth]{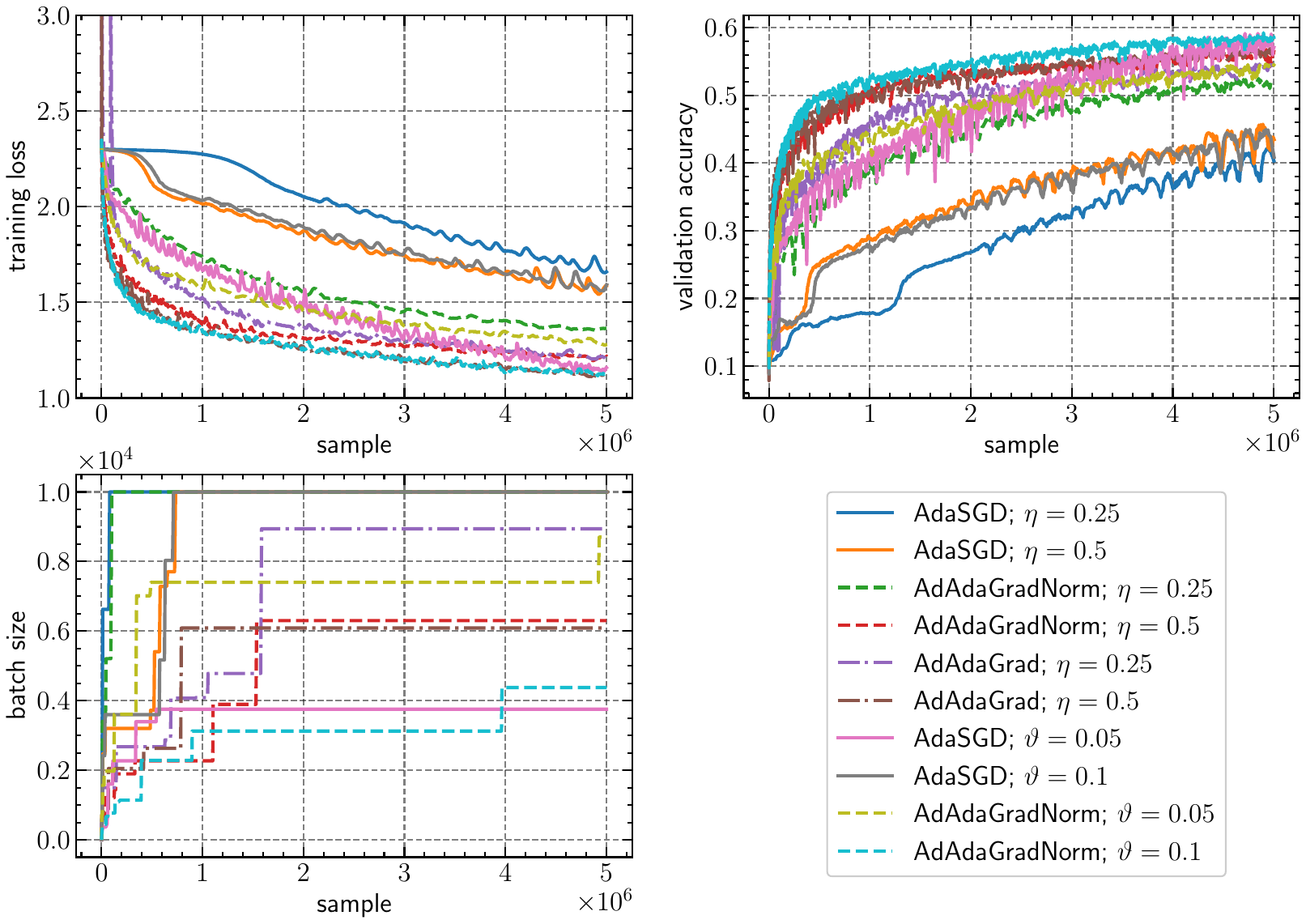}
    \caption{Training loss, validation accuracy, and batch-size curves (vs.~number of training samples) of \AdaSGD, \AdAdaGrad, and \AdAdaGradNorm for a three-layer CNN on the CIFAR-10 dataset.}
        \label{fig:cnn_cifar-10}
    \end{figure}

We next consider a similar problem involving a three-layer CNN with a slightly different architecture on the more challenging CIFAR-10 dataset. We use a training budget of 5 million samples (100 epochs) and a maximum batch size of 10,000 samples. To reduce the computational overhead introduced by the tests, we perform the test every 10 steps.
In \Cref{fig:cnn_cifar-10}, we again observe that \AdaSGD converges more slowly than \AdAdaGrad and \AdAdaGradNorm. This may be due to the lack of a well-crafted learning-rate scaling rule with respect to batch size (cf.~the \emph{scaling rule}) for \SGD in the nonconvex case: the rapid increase in batch size implies a very small effective learning rate, equal to the ratio of the learning rate to the batch size. Without proper rescaling of the learning rate, such a small effective learning rate could slow convergence. We can empirically support this interpretation because \AdaSGD using the inner product test with $\vartheta=0.1$ increases its batch size slowly and eventually plateaus below 4,000. It also converges much faster than its \SGD counterparts, with a final validation accuracy of 57\%, approaching the performance of the adaptive methods. We point out, however, that this instance of \AdaSGD has a much smaller average batch size and therefore requires many more gradient updates than the adaptive methods under an equal budget of training samples (see \Cref{table:cnn_cifar-10}).
    
    \begin{table}[h!]
        \centering
        \caption{Three-layer CNN on CIFAR-10}
        \label{table:cnn_cifar-10}
        \begin{tabular}{ccrrrr}
            \toprule
            Scheme & test & steps & batch size & loss & accuracy \\
            \midrule
            \AdaSGD & $\eta=0.25$ &  523 & 9544 &  1.68 & 0.40  \\
            \AdaSGD & $\eta=0.50$ &  658 & 7592 & 1.59  &  0.43  \\
            \AdAdaGradNorm & $\eta=0.25$ & \textbf{531}  & \textbf{9401} &  \textbf{1.36} &  \textbf{0.52}  \\
            \AdAdaGradNorm & $\eta=0.50$  & \textbf{1261} & \textbf{3964} & \textbf{1.19}  &  \textbf{0.57}  \\
            \AdAdaGrad & $\eta=0.25$ &  903 & 5533 & 1.20 &  0.54  \\
            \AdAdaGrad & $\eta=0.50$  &  \textbf{1123} &  \textbf{4451} & \textbf{1.11}  &  \textbf{0.57}  \\
            \AdaSGD & $\vartheta=0.05$ & \textbf{1597}  & \textbf{3130} & \textbf{1.17}  &  \textbf{0.57}  \\
            \AdaSGD & $\vartheta=0.10$ & 640  & 7806 & 1.64  & 0.41  \\
            \AdAdaGradNorm & $\vartheta=0.05$ & \textbf{780}  & \textbf{6413} &  \textbf{1.29} &  \textbf{0.55}  \\
            \AdAdaGradNorm & $\vartheta=0.10$ & 1948  & 2567 &  1.13 &  0.58  \\
            \bottomrule
        \end{tabular}
    \end{table}

\subsection{\ResNet-18 on CIFAR-10}
We finally train a larger network, \ResNet-18, for image classification on the CIFAR-10 dataset. We use a training budget of 10 million samples (200 epochs) and a maximum batch size of 50,000 samples. Although the focus of this work is on \AdaGrad, we also empirically study the effect of adaptive batch size schemes for \Adam because of its ubiquity in deep learning; we refer to this variant as \AdAdam. The convergence guarantees for \AdAdam are studied in \citet{lau2024adaptive_lm}.

\begin{figure}[t]
    \centering
    \includegraphics[width=\textwidth]{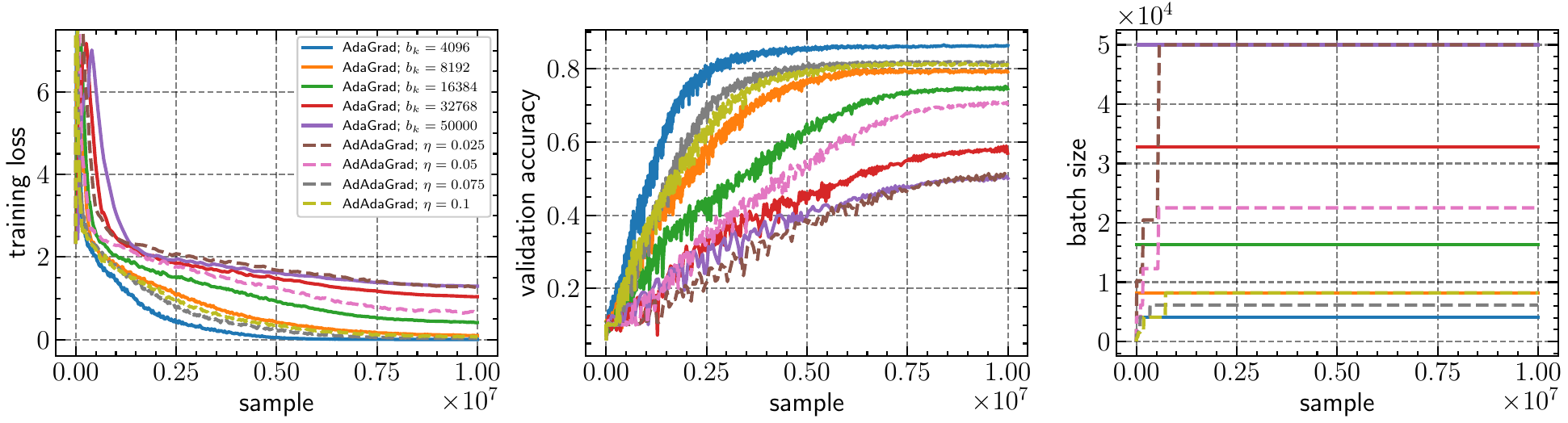}
    \caption{\AdaGrad and \AdAdaGrad for \ResNet-18 on the CIFAR-10 dataset.}
    \label{fig:resnet-18_cifar10_adagrad}
\end{figure}

\paragraph{\AdAdaGrad.}
In \Cref{fig:resnet-18_cifar10_adagrad}, we observe that we need to choose a rather small $\eta$ in order to use full batches during later stages of training for this larger model. Comparing \AdaGrad with a constant batch size of 50,000 and \AdAdaGrad with $\eta=0.025$ (see also \Cref{table:resnet-18_cifar10_full}), \AdAdaGrad is able to use full batches in most of the later stages of training while achieving high accuracy with only 23 additional steps. More generally, our proposed scheme again narrows the generalization gap between smaller and larger constant batch sizes (e.g., the curves for $\eta=0.075$ and $0.1$ lie in the gap between those for constant batch sizes 4096 and 8192).

\begin{figure}[t]
    \centering
    \includegraphics[width=\textwidth]{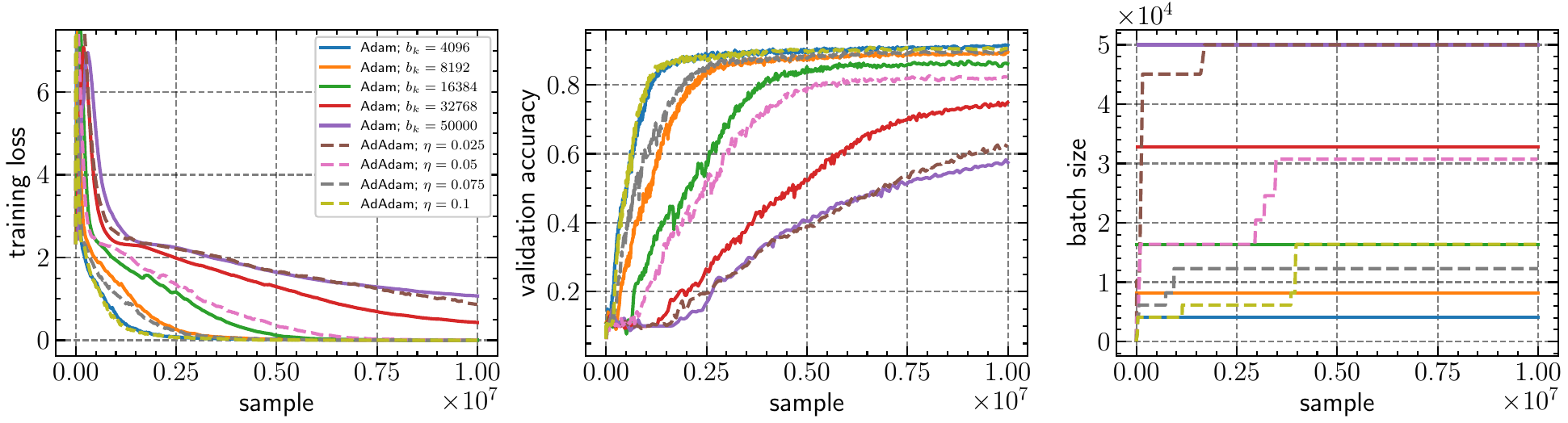}
    \caption{\Adam and \AdAdam for \ResNet-18 on the CIFAR-10 dataset. }
    \label{fig:resnet-18_cifar10_adam}
\end{figure}

\paragraph{\AdAdam.}
In \Cref{fig:resnet-18_cifar10_adam}, we observe a trend for \AdAdam similar to that of \AdAdaGrad, but with faster convergence and larger batch-size increases. It is worth noting from \Cref{table:resnet-18_cifar10_full} that \AdAdam with $\eta=0.1$ and an average batch size of 8880 outperforms \Adam with the smaller constant batch size of 8192 in validation accuracy, while requiring almost 100 fewer steps. This suggests that our proposed scheme may be even more beneficial when combined with \Adam.

\begin{table}[h!]
    \centering
    \caption{\ResNet-18 on CIFAR-10}
    \label{table:resnet-18_cifar10_full}
    \begin{tabular}{ccrrrrr}
        \toprule
        Scheme & test & steps & time (h) & batch size & loss & accuracy \\
        \midrule
        \AdaGrad & N/A & 2441 & 0.88 & 4096 & 0.0042 & 0.8521 \\
        \AdaGrad & N/A & 1220 & 0.70 & 8192 & 0.0808 & 0.8072 \\
        \AdaGrad & N/A & 610 & 0.56 & 16384 & 0.5098 & 0.7264 \\
        \AdaGrad & N/A & 305 & 0.32 & 32768 & 0.9684 & 0.5816 \\
        \AdaGrad & N/A & 199 & 0.23 & 50000 & 1.3625 & 0.4708 \\
        \Adam & N/A & 2441 & 1.20 & 4096 & 0.0003 & 0.9147 \\
        \Adam & N/A & 1220 & 0.97 & 8192 & 0.0004 & 0.8946 \\
        \Adam & N/A & 610 & 0.77 & 16384 & 0.0028 & 0.8628 \\
        \Adam & N/A & 305 & 0.45 & 32768 & 0.4000 & 0.7463 \\
        \Adam & N/A & 199 & 0.33 & 50000 & 1.0680 & 0.5750 \\
        \AdAdaGrad & $\eta=0.025$ & 222 & 0.32 & 44934 & 1.2770 & 0.5107 \\
        \AdAdaGrad & $\eta=0.050$ & 485 & 0.60 & 20615 & 0.6204 & 0.7079 \\
        \AdAdaGrad & $\eta=0.075$ & \textbf{1697} & \textbf{1.02} & \textbf{5892} & \textbf{0.0258} & \textbf{0.8180} \\
        \AdAdaGrad & $\eta=0.100$ & \textbf{1404} & \textbf{0.94} & \textbf{7123} & \textbf{0.0668} & \textbf{0.8085} \\
        \AdAdam & $\eta=0.025$ & 211 & 0.34 & 47380 & 0.9039 & 0.6234 \\
        \AdAdam & $\eta=0.050$ & 426 & 0.60 & 23463 & 0.0061 & 0.8228 \\
        \AdAdam & $\eta=0.075$ & 900 & 0.74 & 11108 & 0.0008 & 0.8983 \\
        \AdAdam & $\eta=0.100$ & \textbf{1126} & \textbf{0.82} & \textbf{8880} & \textbf{0.0000} & \textbf{0.9042} \\
        \bottomrule
    \end{tabular}
\end{table}

\subsection{Discussion}
\label{subsec:discussion}
From the numerical experiments, we can draw several interesting conclusions.
Adaptive batch size schemes are generally optimizer-agnostic, indicating their broad applicability. In particular, coupling adaptive batch sizes with adaptive gradient optimizers can deliver the best of both worlds: we can narrow the generalization gap while still benefiting from the faster convergence of adaptive gradient optimizers. The computational overhead introduced by the batch-size tests could limit the practical use of the proposed methods in large-scale applications; further engineering effort, as well as development for distributed training, is necessary to fully realize the potential benefits of the proposed adaptive batch size schemes.

\section{Concluding Remarks}
\label{sec:conclusion}
In this work, we demonstrate the versatility of adaptive sampling methods as generic adaptive batch size schemes for adaptive gradient optimizers, supported by both convergence guarantees and numerical results. This opens up several promising research directions for adaptive batch size schemes in large-scale model training. On the theoretical side, it would be interesting to study the convergence guarantees of this class of methods when combined with other stochastic gradient optimizers, such as momentum-based methods, as well as proximal \SGD methods for constrained problems with deterministic nonsmooth regularizers. On the practical side, exploring the implementation of adaptive batch size schemes under various parallelism paradigms for large-scale distributed training---including data, tensor, and pipeline parallelism \citep{shoeybi2019megatron, rajbhandari2020zero, smith2022using, zhao2023pytorch}---is worthwhile; see, e.g., \citet{lau2024adaptive_lm} for data and model parallelism. This line of work aims to ensure that these schemes are viable for large-scale applications such as the (pre-)training of autoregressive language and image models. Furthermore, examining the impact of adaptive batch size schemes for adaptive gradient methods, in contrast to those designed for \SGD, particularly for transformer-based language models in addition to the CNN-based vision tasks discussed in this paper, is important. Unlike the marginal utility of adaptive methods for CNNs and RNNs \citep{wilson2017marginal}, adaptive gradient methods such as \Adam significantly outperform \SGD when optimizing transformers \citep{zhang2020why, zhang2020adaptive, jiang2022does, kunstner2023noise, pan2023toward, ahn2023linear}.

\subsection*{Code Availability}
All analysis code is available in the GitHub repository: \url{https://github.com/timlautk/AdAdaGrad}.

\section*{Acknowledgments}
The research of Han Liu is supported by U.S. National Institutes of Health (NIH) (Grant No. R01LM01372201), U.S. National Science Foundation (NSF) (Grant Nos. RI 1840857, TRIPOD 1740735, DMS 1454377-CAREER, and IIS 1546482, and Alfred P. Sloan Fellowship. The research of Mladen Kolar is supported in part by NSF (Grant No. ECCS-2216912). This research is supported in part by computational resources and staff contributions provided by the Data Science Institute at the University of Chicago, through its AI + Science Research Funding Initiatives. Part of the work of Tim Tsz-Kit Lau and Mladen Kolar was performed while they were at the University of Chicago Booth School of Business.

\appendix
\titleformat{\section}{\normalfont\Large\bfseries\sffamily}{Appendix \thesection.}{0.5em}{}

\section{Proofs of Main Text}
\label{sec:proofs}
We provide the omitted proofs of the main text in this section.


\subsection{Preparatory Definitions, Propositions and Lemmas}
We give various additional technical definitions, propositions and lemmas before giving the proofs of the theorems.

\subsubsection{Formal Statements Corresponding to \Cref{prop:tests}}
We now state precise versions of \Cref{prop:tests}.

\begin{proposition}[Exact variance norm test]
\label{prop:norm}
Suppose that, for every iteration $k\in\NN^*$, the batch gradient $\nabla F_{\calB_k}(x_k)$ is conditionally unbiased, that is,
\[
\Ex_k[\nabla F_{\calB_k}(x_k)] = \nabla F(x_k),
\]
and the exact variance norm test \eqref{eqn:exact_norm} holds with some constant $\eta>0$:
\[
\Ex_k\left[\|\nabla F_{\calB_k}(x_k) - \nabla F(x_k)\|^2\right]
\le \eta^2 \|\nabla F(x_k)\|^2.
\]
Then
\[
\Ex_k\left[\|\nabla F_{\calB_k}(x_k)\|^2\right]
\le (1+\eta^2)\|\nabla F(x_k)\|^2.
\]
\end{proposition}

\begin{proof}
Using the conditional unbiasedness of $\nabla F_{\calB_k}(x_k)$, we have
\begin{align*}
\Ex_k\left[\|\nabla F_{\calB_k}(x_k) - \nabla F(x_k)\|^2\right]
&= \Ex_k\left[\|\nabla F_{\calB_k}(x_k)\|^2\right]
   - 2\dotp{\Ex_k[\nabla F_{\calB_k}(x_k)]}{\nabla F(x_k)}
   + \|\nabla F(x_k)\|^2 \\
&= \Ex_k\left[\|\nabla F_{\calB_k}(x_k)\|^2\right]
   - \|\nabla F(x_k)\|^2.
\end{align*}
Combining this identity with \eqref{eqn:exact_norm} yields the claim.
\end{proof}

\begin{proposition}[Exact variance inner product test and orthogonality test]
\label{prop:inner_prod}
Suppose that, for every iteration $k\in\NN^*$ with $\nabla F(x_k)\neq 0$, the samples in $\calB_k$ are conditionally i.i.d., and the exact variance inner product test \eqref{eqn:exact_inner_prod} and exact variance orthogonality test \eqref{eqn:exact_ortho} hold with constants $\vartheta>0$ and $\nu>0$, respectively. Then
\[
\Ex_k\left[\|\nabla F_{\calB_k}(x_k)\|^2\right]
\le (1+\vartheta^2+\nu^2)\|\nabla F(x_k)\|^2.
\]
\end{proposition}

\begin{proof}
This is part of the result of \citet{bollapragada2018adaptive}, Lemma 3.1. We include its proof here with our notation for completeness. 

Write the batch gradient as
\[
\nabla F_{\calB_k}(x_k)
= \frac{1}{b_k}\sum_{r=1}^{b_k} \nabla f(x_k;\xi_{k,r}),
\]
where $\xi_{k,1},\ldots,\xi_{k,b_k}$ are the conditionally i.i.d.~samples in $\calB_k$.
For $r=1,\ldots,b_k$, define
\[
a_{k,r}
\coloneqq \dotp{\nabla f(x_k;\xi_{k,r})}{\nabla F(x_k)} - \|\nabla F(x_k)\|^2
\]
and
\[
u_{k,r}
\coloneqq \nabla f(x_k;\xi_{k,r})
- \frac{\dotp{\nabla f(x_k;\xi_{k,r})}{\nabla F(x_k)}}{\|\nabla F(x_k)\|^2}\nabla F(x_k).
\]
Since the samples are conditionally i.i.d.~and
\[
\Ex_k[\nabla f(x_k;\xi_{k,r})] = \nabla F(x_k),
\]
we have $\Ex_k[a_{k,r}]=0$ and $\Ex_k[u_{k,r}]=0$. Hence
\[
\dotp{\nabla F_{\calB_k}(x_k)}{\nabla F(x_k)} - \|\nabla F(x_k)\|^2
= \frac{1}{b_k}\sum_{r=1}^{b_k} a_{k,r},
\]
and
\[
\nabla F_{\calB_k}(x_k)
- \frac{\dotp{\nabla F_{\calB_k}(x_k)}{\nabla F(x_k)}}{\|\nabla F(x_k)\|^2}\nabla F(x_k)
= \frac{1}{b_k}\sum_{r=1}^{b_k} u_{k,r}.
\]
Therefore, by conditional independence and centering,
\begin{align*}
\Ex_k\left[\left(\dotp{\nabla F_{\calB_k}(x_k)}{\nabla F(x_k)} - \|\nabla F(x_k)\|^2\right)^2\right]
&= \Ex_k\left[\left(\frac{1}{b_k}\sum_{r=1}^{b_k} a_{k,r}\right)^{\negthickspace\negthinspace2}\right] \\
&= \frac{1}{b_k^2}\sum_{r=1}^{b_k}\Ex_k[a_{k,r}^2] \\
&= \frac{1}{b_k}\Ex_k[a_{k,1}^2] \\
&\le \vartheta^2 \|\nabla F(x_k)\|^4,
\end{align*}
and similarly
\begin{align*}
\Ex_k\left[\left\|\nabla F_{\calB_k}(x_k)
- \frac{\dotp{\nabla F_{\calB_k}(x_k)}{\nabla F(x_k)}}{\|\nabla F(x_k)\|^2}\nabla F(x_k)\right\|^2\right]
&= \Ex_k\left[\left\|\frac{1}{b_k}\sum_{r=1}^{b_k} u_{k,r}\right\|^2\right] \\
&= \frac{1}{b_k^2}\sum_{r=1}^{b_k}\Ex_k[\|u_{k,r}\|^2] \\
&= \frac{1}{b_k}\Ex_k[\|u_{k,1}\|^2] \\
&\le \nu^2 \|\nabla F(x_k)\|^2.
\end{align*}

Now decompose $\nabla F_{\calB_k}(x_k)$ into its components parallel and orthogonal to $\nabla F(x_k)$:
\[
\nabla F_{\calB_k}(x_k)
= \frac{\dotp{\nabla F_{\calB_k}(x_k)}{\nabla F(x_k)}}{\|\nabla F(x_k)\|^2}\nabla F(x_k) + r_k,
\]
where
\[
r_k
\coloneqq \nabla F_{\calB_k}(x_k)
- \frac{\dotp{\nabla F_{\calB_k}(x_k)}{\nabla F(x_k)}}{\|\nabla F(x_k)\|^2}\nabla F(x_k).
\]
Since the two components are orthogonal,
\[
\|\nabla F_{\calB_k}(x_k)\|^2
= \frac{\left(\dotp{\nabla F_{\calB_k}(x_k)}{\nabla F(x_k)}\right)^2}{\|\nabla F(x_k)\|^2}
+ \|r_k\|^2.
\]
Taking conditional expectation and using
\[
\Ex_k\left[\dotp{\nabla F_{\calB_k}(x_k)}{\nabla F(x_k)}\right]
= \|\nabla F(x_k)\|^2,
\]
we obtain
\begin{align*}
\Ex_k\left[\|\nabla F_{\calB_k}(x_k)\|^2\right]
&= \frac{\Ex_k\left[\left(\dotp{\nabla F_{\calB_k}(x_k)}{\nabla F(x_k)}\right)^2\right]}{\|\nabla F(x_k)\|^2}
   + \Ex_k[\|r_k\|^2] \\
&= \frac{\Ex_k\left[\left(\dotp{\nabla F_{\calB_k}(x_k)}{\nabla F(x_k)} - \|\nabla F(x_k)\|^2\right)^2\right]
   + \|\nabla F(x_k)\|^4}{\|\nabla F(x_k)\|^2}
   + \Ex_k[\|r_k\|^2] \\
&\le (1+\vartheta^2+\nu^2)\|\nabla F(x_k)\|^2.
\end{align*}
\end{proof}

\subsubsection{Technical Lemmas}
We first record a simple summation lemma for nonnegative sequences.

\begin{lemma}\label{lem:seq}
    Let $(a_k)_{k\in\NN}\subset\Rp$ with $a_0>0$, and define $s_k\coloneqq\sum_{i=0}^k a_i$ for each $k\in\NN$. Then
    \begin{align*}
        \sumK \frac{a_k}{s_k^{3/2}} &\le \frac{2}{\sqrt{a_0}}, \\
        \sumK \frac{a_k}{s_k} &\le \log s_K - \log a_0.
    \end{align*}
\end{lemma}    
    
\begin{proof}
    For each $k\in\setK$, we have $a_k = s_k - s_{k-1}$. Since the function $t\mapsto t^{-3/2}$ is decreasing on $\Rpp$,
    \[
        \frac{a_k}{s_k^{3/2}}
        = \frac{s_k-s_{k-1}}{s_k^{3/2}}
        \le \int_{s_{k-1}}^{s_k} t^{-3/2}\,\diff t.
    \]
    Summing over $k=1,\ldots,K$ yields
    \[
        \sumK \frac{a_k}{s_k^{3/2}}
        \le \int_{a_0}^{s_K} t^{-3/2}\,\diff t
        = \frac{2}{\sqrt{a_0}} - \frac{2}{\sqrt{s_K}}
        \le \frac{2}{\sqrt{a_0}}.
    \]
    Similarly, since $t\mapsto t^{-1}$ is decreasing on $\Rpp$,
    \[
        \frac{a_k}{s_k}
        = \frac{s_k-s_{k-1}}{s_k}
        \le \int_{s_{k-1}}^{s_k} t^{-1}\,\diff t.
    \]
    Summing over $k=1,\ldots,K$ gives
    \[
        \sumK \frac{a_k}{s_k}
        \le \int_{a_0}^{s_K} t^{-1}\,\diff t
        = \log s_K - \log a_0.
    \]
\end{proof}

We also state without proof the descent lemma for $(L_0, L_1)$-smooth functions; see \citet{zhang2020improved}, Lemma A.3.

\begin{lemma}[Descent lemma for $(L_0, L_1)$-smooth functions]
\label{lem:thm_AdaGrad-Norm_gen_smooth}
    If \Cref{ass:gen_smooth} holds, then for any $x,y\in\RR^d$ satisfying $L_1\norm{x-y}\le 1$,
    \[
        F(y) \le F(x) + \dotp{\nabla F(x)}{y - x} + \frac{L_0 + L_1\norm{\nabla F(x)}}{2}\norm{x-y}^2.
    \]
\end{lemma}

\subsection{Convergence Results for \AdaGradNorm and \AdaGrad}

\subsubsection{Proof of \Cref{thm:AdaGrad-Norm}}
\label{subsubsec:proof_thm6}

We start by providing two results.

\begin{lemma}\label{lem:thm_AdaGrad-Norm}
    Define
    \[
        \varphi_0 \coloneqq \frac{\|\nabla F(x_1)\|^2}{\sqrt{v_0}},
        \qquad
        (\forall k\in\NN^*)\quad
        \varphi_k \coloneqq \frac{\|\nabla F(x_k)\|^2}{\sqrt{v_k}},
    \]
    and adopt the convention $g_0\coloneqq 0$. Then, for any $\rho>0$ and every $k\in\NN^*$,
    \begin{multline}
        \lrdotp{\gradFxk}{\Ex_k\left[\left( \frac{1}{\sqrt{\vkmo}} - \frac{1}{\sqrt{v_k}}\right) g_k\right]} \\
        \le \frac12\left(1+\frac1\rho\right)\frac{\norm{\gradFxk}^2}{\sqrt{\vkmo}}
        + \frac{\tau}{2}\Ex_k[\varphi_{k-1} - \varphi_k]
        + \frac{\tau L^2\alpha^2}{2}(1+\rho\tau)\frac{\norm{\gkmo}^2}{\vkmo^{\sfrac32}}.
    \end{multline}
\end{lemma}

\begin{proof}[Proof of \Cref{lem:thm_AdaGrad-Norm}]
Note that
\begin{equation}\label{eqn:vk}
        \frac{1}{\sqrt{\vkmo}} - \frac{1}{\sqrt{v_k}}
        = \frac{v_k - \vkmo}{\sqrt{\vkmo}\sqrt{v_k}(\sqrt{v_k} + \sqrt{\vkmo})}
        = \frac{\norm{g_k}^2}{\sqrt{\vkmo}\sqrt{v_k}(\sqrt{v_k} + \sqrt{\vkmo})}.
\end{equation}
Hence
\begin{align*}
        &\lrdotp{\gradFxk}{\Ex_k\left[\left( \frac{1}{\sqrt{\vkmo}} - \frac{1}{\sqrt{v_k}}\right) g_k\right]} \\
        &\le \frac{\norm{\gradFxk}}{\sqrt{\vkmo}}
        \Ex_k\left[\frac{\norm{g_k}^3}{\sqrt{v_k}(\sqrt{v_k} + \sqrt{\vkmo})}\right] \\
        &\le \frac{\norm{\gradFxk}}{\sqrt{\vkmo}}
        \Ex_k\left[\frac{\norm{g_k}^2}{\sqrt{v_k} + \sqrt{\vkmo}}\right]
        && \text{since } v_k \ge \norm{g_k}^2 \\
        &\le \frac{\norm{\gradFxk}^2}{2\sqrt{\vkmo}}
        + \frac{1}{2\sqrt{\vkmo}}
        \left(\Ex_k\left[\frac{\norm{g_k}^2}{\sqrt{v_k} + \sqrt{\vkmo}}\right]\right)^2 \\
        &\le \frac{\norm{\gradFxk}^2}{2\sqrt{\vkmo}}
        + \frac{1}{2\sqrt{\vkmo}}\Ex_k[\norm{g_k}^2]
        \Ex_k\left[\frac{\norm{g_k}^2}{(\sqrt{v_k} + \sqrt{\vkmo})^2}\right] \\
        &\le \frac{\norm{\gradFxk}^2}{2\sqrt{\vkmo}}
        + \frac{\tau\norm{\gradFxk}^2}{2\sqrt{\vkmo}}
        \Ex_k\left[\frac{\norm{g_k}^2}{(\sqrt{v_k} + \sqrt{\vkmo})^2}\right].
\end{align*}
Moreover, by \eqref{eqn:vk},
\[
        \frac{\norm{g_k}^2}{\sqrt{\vkmo}(\sqrt{v_k} + \sqrt{\vkmo})^2}
        \le \frac{\norm{g_k}^2}{\sqrt{\vkmo}\sqrt{v_k}(\sqrt{v_k} + \sqrt{\vkmo})}
        = \frac{1}{\sqrt{\vkmo}} - \frac{1}{\sqrt{v_k}}.
\]
Therefore,
\begin{equation}
\label{eqn:lem_AdaGrad-Norm_aux}
        \lrdotp{\gradFxk}{\Ex_k\left[\left( \frac{1}{\sqrt{\vkmo}} - \frac{1}{\sqrt{v_k}}\right) g_k\right]}
        \le \frac{\norm{\gradFxk}^2}{2\sqrt{\vkmo}}
        + \frac{\tau}{2}\norm{\gradFxk}^2
        \Ex_k\left[\frac{1}{\sqrt{\vkmo}} - \frac{1}{\sqrt{v_k}}\right].
\end{equation}    
For $k=1$, the claim follows immediately from \eqref{eqn:lem_AdaGrad-Norm_aux}, the definition of $\varphi_0$, the inequality $\frac12 \le \frac12(1+\rho^{-1})$, and the convention $g_0=0$. Assume now that $k\ge 2$.

We decompose
\begin{multline*}
    \norm{\gradFxk}^2 \Ex_k\left[\frac{1}{\sqrt{\vkmo}} - \frac{1}{\sqrt{v_k}}\right] \\
    = \Ex_k\left[\frac{\norm{\nabla F(\xkmo)}^2}{\sqrt{\vkmo}} - \frac{\norm{\gradFxk}^2}{\sqrt{v_k}}\right]
    + \frac{\norm{\gradFxk}^2 - \norm{\nabla F(\xkmo)}^2}{\sqrt{\vkmo}}.
\end{multline*}
By the reverse triangle inequality and \Cref{ass:grad_Lip},
\[
    \norm{\gradFxk} - \norm{\nabla F(\xkmo)}
    \le \norm{\gradFxk - \nabla F(\xkmo)}
    \le L\norm{x_k - \xkmo},
\]
and by the triangle inequality,
\[
    \norm{\nabla F(\xkmo)}
    \le \norm{\gradFxk} + L\norm{x_k - \xkmo}.
\]
Hence
\begin{align*}
    \norm{\gradFxk}^2 - \norm{\nabla F(\xkmo)}^2
    &= \left(\norm{\gradFxk} - \norm{\nabla F(\xkmo)}\right)
       \left(\norm{\gradFxk} + \norm{\nabla F(\xkmo)}\right) \\
    &\le L^2\norm{x_k - \xkmo}^2 + 2L\norm{\gradFxk}\norm{x_k - \xkmo}.
\end{align*}
Since $x_k - \xkmo = -\alpha \gkmo/\sqrt{\vkmo}$, we obtain
\begin{multline*}
    \norm{\gradFxk}^2 \Ex_k\left[\frac{1}{\sqrt{\vkmo}} - \frac{1}{\sqrt{v_k}}\right] \\
    \le \Ex_k[\varphi_{k-1} - \varphi_k]
    + \frac{L^2\alpha^2}{\vkmo^{\sfrac32}}\norm{\gkmo}^2
    + \frac{2L\alpha}{\vkmo}\norm{\gradFxk}\norm{\gkmo}.
\end{multline*}
Applying Young's inequality with parameter $\rho>0$,
\[
    \frac{\tau}{2}\cdot \frac{2L\alpha}{\vkmo}\norm{\gradFxk}\norm{\gkmo}
    \le \frac{1}{2\rho}\frac{\norm{\gradFxk}^2}{\sqrt{\vkmo}}
    + \frac{\rho\tau^2L^2\alpha^2}{2}\frac{\norm{\gkmo}^2}{\vkmo^{\sfrac32}}.
\]
Substituting this bound into \eqref{eqn:lem_AdaGrad-Norm_aux} yields
\begin{align*}
    &\quad \lrdotp{\gradFxk}{\Ex_k\left[\left( \frac{1}{\sqrt{\vkmo}} - \frac{1}{\sqrt{v_k}}\right) g_k\right]} \\
    &\le \frac{\norm{\gradFxk}^2}{2\sqrt{\vkmo}}
    + \frac{\tau}{2}\Ex_k[\varphi_{k-1} - \varphi_k]
    + \frac{\tau L^2\alpha^2}{2}\frac{\norm{\gkmo}^2}{\vkmo^{\sfrac32}}
    + \frac{1}{2\rho}\frac{\norm{\gradFxk}^2}{\sqrt{\vkmo}}
    + \frac{\rho\tau^2L^2\alpha^2}{2}\frac{\norm{\gkmo}^2}{\vkmo^{\sfrac32}} \\
    &= \frac12\left(1+\frac1\rho\right)\frac{\norm{\gradFxk}^2}{\sqrt{\vkmo}}
    + \frac{\tau}{2}\Ex_k[\varphi_{k-1} - \varphi_k]
    + \frac{\tau L^2\alpha^2}{2}(1+\rho\tau)\frac{\norm{\gkmo}^2}{\vkmo^{\sfrac32}},
\end{align*}
which completes the proof.
\end{proof}  

\begin{lemma}\label{lem:thm_AdaGrad-Norm_2}
    For any positive constants $(a,b)\in\Rpp^2$, if $x>0$ satisfies
    $x\le a+b\log x$, then
    $x\le 2a - 2b + 4b\log(b/2 +1) \le 2a + 4b\log(b/2+1)$.
\end{lemma}
\begin{proof}[Proof of \Cref{lem:thm_AdaGrad-Norm_2}]
    Let $h(t) \coloneqq t/2 - b\log t$ for $t>0$. Then
    \[
        h'(t) = 1/2 - b/t,
        \qquad
        h''(t) = b/t^2 > 0.
    \]
    Hence $h$ is minimized at $t=2b$, so
    \[
        h(t) \ge h(2b) = b - b\log(2b).
    \]
    Applying this with $t=x$, we obtain
    \[
        x - b\log x
        = x/2 + \left(x/2 - b\log x\right)
        \ge x/2 + b - b\log(2b).
    \]
    Since $x\le a+b\log x$, equivalently $x-b\log x\le a$, it follows that
    \[
        x/2 + b - b\log(2b) \le a,
    \]
    and therefore
    \[
        x \le 2a - 2b + 2b\log(2b).
    \]

    By the A.M.-G.M. inequality, $b+2 \ge 2\sqrt{2b}$, and taking logarithms gives
    \[
        \log(b+2) \ge \log 2 + \frac12(\log 2 + \log b).
    \]
    Therefore,
    \[
        b\log(2b)
        = b(\log 2 + \log b)
        \le 2b(\log(b+2) - \log 2)
        = 2b\log\left(b/2 + 1\right).
    \]
    Substituting this bound into the previous inequality yields
    \[
        x\le 2a - 2b + 4b\log\left(b/2 + 1\right),
    \]
    and the final inequality is immediate.
\end{proof}

If $\nabla F(x_k)=0$ for some $k\in\setK$, then the conclusion is immediate. Hence, when the augmented inner product test is used, we may assume $\nabla F(x_k)\neq 0$ for all $k\in\setK$. Under the corresponding hypotheses of \Cref{prop:norm,prop:inner_prod}, for every $k\in\NN^*$,
\begin{equation}
\label{eqn:thm_AdaGrad-Norm_esg}
    \Ex_k[\|g_k\|^2] \le \tau \|\nabla F(x_k)\|^2.
\end{equation}
By \Cref{ass:grad_Lip},
\begin{align*}
    F(x_{k+1})
    &\le F(x_k) + \dotp{\gradFxk}{x_{k+1}-x_k} + \frac{L}{2}\|x_{k+1}-x_k\|^2 \\
    &= F(x_k) - \alpha \dotp{\gradFxk}{\frac{g_k}{\sqrt{v_k}}}
       + \frac{L\alpha^2}{2}\frac{\|g_k\|^2}{v_k}.
\end{align*}
Taking conditional expectation with respect to $\calF_k$ gives
\begin{equation*}
    \Ex_k[F(\xkpo)]
    \le F(x_k) - \alpha \lrdotp{\gradFxk}{\Ex_k\left[\frac{g_k}{\sqrt{v_k}}\right]}
    + \frac{L\alpha^2}{2}\Ex_k\left[\frac{\|g_k\|^2}{v_k}\right].
\end{equation*}
Moreover,
\begin{align}
    \lrdotp{\gradFxk}{\Ex_k\left[\frac{g_k}{\sqrt{v_k}}\right]}
    &= \lrdotp{\gradFxk}{\Ex_k\left[\frac{g_k}{\sqrt{\vkmo}}\right]}
       + \lrdotp{\gradFxk}{\Ex_k\left[\left(\frac{1}{\sqrt{v_k}}-\frac{1}{\sqrt{\vkmo}}\right)g_k\right]} \nonumber\\
    &= \frac{\|\gradFxk\|^2}{\sqrt{\vkmo}}
       + \lrdotp{\gradFxk}{\Ex_k\left[\left(\frac{1}{\sqrt{v_k}}-\frac{1}{\sqrt{\vkmo}}\right)g_k\right]}.
    \label{eqn:thm_AdaGrad-Norm_3}
\end{align}
Hence
\begin{multline}
\label{eqn:thm_AdaGrad-Norm_1}
    \Ex_k[F(\xkpo)]
    \le F(x_k) - \alpha \frac{\|\gradFxk\|^2}{\sqrt{\vkmo}} \\
    + \alpha \lrdotp{\gradFxk}{\Ex_k\left[\left(\frac{1}{\sqrt{\vkmo}}-\frac{1}{\sqrt{v_k}}\right)g_k\right]}
    + \frac{L\alpha^2}{2}\Ex_k\left[\frac{\|g_k\|^2}{v_k}\right].
\end{multline}
Combining \eqref{eqn:thm_AdaGrad-Norm_1} with \Cref{lem:thm_AdaGrad-Norm} and taking total expectation, we obtain
\begin{multline*}
    \Ex[F(\xkpo)]
    \le \Ex[F(x_k)]
    - \frac{\alpha}{2}\left(1-\frac{1}{\rho}\right)\Ex\left[\frac{\|\gradFxk\|^2}{\sqrt{\vkmo}}\right] \\
    + \frac{\tau\alpha}{2}\Ex[\varphi_{k-1}-\varphi_k]
    + \frac{L\alpha^2}{2}\Ex\left[\frac{\|g_k\|^2}{v_k}\right]
    + \frac{\tau L^2\alpha^3}{2}(1+\rho\tau)\Ex\left[\frac{\|g_{k-1}\|^2}{\vkmo^{3/2}}\right].
\end{multline*}
Now define
\[
    A_K \coloneqq \sum_{k=1}^K \Ex\left[\frac{\|\gradFxk\|^2}{\sqrt{\vkmo}}\right].
\]
Summing the previous inequality over $k=1,\ldots,K$, using $F(x_{K+1})\ge F^\star$, $\varphi_K\ge 0$, $g_0=0$, and \Cref{lem:seq}, yields
\begin{align*}
\frac{\alpha}{2}\Big(1&-\frac{1}{\rho}\Big) A_K
\\
&\le F(x_1)-F^\star + \frac{\tau\alpha}{2}\varphi_0
+ \frac{L\alpha^2}{2}\left(\Ex[\log v_K]-\log v_0\right) 
+ \frac{\tau L^2\alpha^3}{2}(1+\rho\tau)\sum_{k=1}^K \Ex\left[\frac{\|g_{k-1}\|^2}{\vkmo^{3/2}}\right] \\
&\le F(x_1)-F^\star + \frac{\tau\alpha}{2}\frac{\|\nabla F(x_1)\|^2}{\sqrt{v_0}}
+ \frac{L\alpha^2}{2}\left(\Ex[\log v_K]-\log v_0\right)
+ \tau L^2\alpha^3(1+\rho\tau)\frac{1}{\sqrt{v_0}}.
\end{align*}
Therefore,
\[
    A_K
    \le c_1 + \frac{L\alpha}{1-\rho^{-1}}\Ex[\log v_K]
    \le c_1^+ + c_2\Ex[\log v_K].
\]

To control $\Ex[\sqrt{v_K}]$, observe that
\[
    \sqrt{v_K}
    = \sqrt{v_0} + \sum_{k=1}^K \frac{\|g_k\|^2}{\sqrt{v_k}+\sqrt{v_{k-1}}}
    \le \sqrt{v_0} + \sum_{k=1}^K \frac{\|g_k\|^2}{\sqrt{v_{k-1}}}.
\]
Taking expectation and using \eqref{eqn:thm_AdaGrad-Norm_esg}, we get
\[
    x\coloneqq \Ex[\sqrt{v_K}]
    \le \sqrt{v_0} + \tau A_K
    \le \sqrt{v_0} + \tau c_1^+ + \tau c_2\Ex[\log v_K].
\]
Since $\log v_K = 2\log \sqrt{v_K}$ and $\log$ is concave,
\[
    \Ex[\log v_K] \le 2\log \Ex[\sqrt{v_K}] = 2\log x.
\]
Thus
\[
    x \le \sqrt{v_0} + \tau c_1^+ + 2\tau c_2\log x.
\]
Applying \Cref{lem:thm_AdaGrad-Norm_2} with $a=\sqrt{v_0}+\tau c_1^+$ and $b=2\tau c_2$, we obtain
\[
    \Ex[\sqrt{v_K}] \le c_3.
\]
Consequently,
\[
    A_K \le c_1^+ + 2c_2\log c_3.
\]

Now set
\[
    S_K \coloneqq \sum_{k=1}^K \|\nabla F(x_k)\|^2.
\]
Since $v_{k-1}\le v_K$ for all $k\in\setK$,
\[
    A_K \ge \Ex\left[\frac{S_K}{\sqrt{v_K}}\right].
\]
By Cauchy-Schwarz,
\[
    \Ex[\sqrt{S_K}]^2
    \le \Ex\left[\frac{S_K}{\sqrt{v_K}}\right]\Ex[\sqrt{v_K}]
    \le A_K \Ex[\sqrt{v_K}]
 \le c_3(c_1^+ + 2c_2\log c_3).
\]
Therefore, for any $\varepsilon>0$,
\[
    \Prob\left(\min_{k\in\setK}\|\nabla F(x_k)\|^2 > \varepsilon\right)
    \le \Prob(S_K > K\varepsilon)
    = \Prob(\sqrt{S_K} > \sqrt{K\varepsilon})
    \le \frac{\Ex[\sqrt{S_K}]}{\sqrt{K\varepsilon}}.
\]
Choosing
\[
    \varepsilon
    = \frac{c_3(c_1^+ + 2c_2\log c_3)}{K\delta^2}
\]
gives
\[
    \Prob\left(\min_{k\in\setK}\|\nabla F(x_k)\|^2 > \frac{c_3(c_1^+ + 2c_2\log c_3)}{K\delta^2}\right)
    \le \delta,
\]
which completes the proof.

\subsubsection{Proof of \Cref{thm:AdaGrad}}    
The proof follows the same strategy as that of \Cref{thm:AdaGrad-Norm}, but the estimates must be carried out coordinate-wise. Write $g_{k,j}\coloneqq[g_k]_j$, $v_{k,j}\coloneqq[v_k]_j$, and set
\[
\tau \coloneqq 1+\eta^2, \qquad
H_k \coloneqq \left\|\frac{1}{\sqrt{v_k}}\odot g_k\right\|^2, \qquad
\Gamma_k \coloneqq \sum_{j=1}^d \frac{1}{\sqrt{v_{k,j}}},
\]
\[
\tvarphi_{0,j}\coloneqq\frac{(\partial_jF(x_1))^2}{\sqrt{v_{0,j}}}, \qquad
(\forall k\in\NN^*)\quad
\tvarphi_{k,j}\coloneqq\frac{(\partial_jF(x_k))^2}{\sqrt{v_{k,j}}}, \qquad
j\in\setd.
\]
We also set $H_0\coloneqq 0$.

Since $g_k=\nabla F_{\calB_k}(x_k)$ is conditionally unbiased, the coordinate-wise exact variance norm test \eqref{eqn:co_exact_norm} implies
\[
\Ex_k[g_{k,j}^2]
=
\Ex_k[(g_{k,j}-\partial_jF(x_k))^2] + (\partial_jF(x_k))^2
\le \tau (\partial_jF(x_k))^2
\]
for every $(k,j)\in\NN^*\times\setd$.

By \Cref{ass:grad_Lip} and the AdaGrad update $x_{k+1}=x_k-\alpha g_k\odot v_k^{-\half}$,
\begin{align*}
\Ex_k[F(x_{k+1})]
&\le F(x_k) - \alpha \left\langle \nabla F(x_k), \Ex_k\left[\frac{1}{\sqrt{v_k}}\odot g_k\right]\right\rangle
   + \frac{L\alpha^2}{2}\Ex_k[H_k] \\
&= F(x_k)
   - \alpha \sum_{j=1}^d \frac{(\partial_jF(x_k))^2}{\sqrt{v_{k-1,j}}}
   + \alpha \sum_{j=1}^d T_{k,j}
   + \frac{L\alpha^2}{2}\Ex_k[H_k],
\end{align*}
where
\[
T_{k,j}\coloneqq
\partial_jF(x_k)\,
\Ex_k\left[\left(\frac{1}{\sqrt{v_{k-1,j}}}-\frac{1}{\sqrt{v_{k,j}}}\right)g_{k,j}\right].
\]

Since $v_{k,j}=v_{k-1,j}+g_{k,j}^2$,
\[
\frac{1}{\sqrt{v_{k-1,j}}}-\frac{1}{\sqrt{v_{k,j}}}
=
\frac{g_{k,j}^2}{\sqrt{v_{k-1,j}}\sqrt{v_{k,j}}(\sqrt{v_{k,j}}+\sqrt{v_{k-1,j}})}.
\]
Using $\sqrt{v_{k,j}}\ge |g_{k,j}|$, Young's inequality, and Cauchy-Schwarz, we obtain
\begin{align*}
T_{k,j}
&\le \frac{|\partial_jF(x_k)|}{\sqrt{v_{k-1,j}}}
\Ex_k\left[\frac{|g_{k,j}|^3}{\sqrt{v_{k,j}}(\sqrt{v_{k,j}}+\sqrt{v_{k-1,j}})}\right] \\
&\le \frac{|\partial_jF(x_k)|}{\sqrt{v_{k-1,j}}}
\Ex_k\left[\frac{g_{k,j}^2}{\sqrt{v_{k,j}}+\sqrt{v_{k-1,j}}}\right] \\
&\le \frac{(\partial_jF(x_k))^2}{2\sqrt{v_{k-1,j}}}
   + \frac{1}{2\sqrt{v_{k-1,j}}}
     \left(\Ex_k\left[\frac{g_{k,j}^2}{\sqrt{v_{k,j}}+\sqrt{v_{k-1,j}}}\right]\right)^2 \\
&\le \frac{(\partial_jF(x_k))^2}{2\sqrt{v_{k-1,j}}}
   + \frac{\Ex_k[g_{k,j}^2]}{2\sqrt{v_{k-1,j}}}
     \Ex_k\left[\frac{g_{k,j}^2}{(\sqrt{v_{k,j}}+\sqrt{v_{k-1,j}})^2}\right] \\
&\le \frac{(\partial_jF(x_k))^2}{2\sqrt{v_{k-1,j}}}
   + \frac{\tau}{2}(\partial_jF(x_k))^2
     \Ex_k\left[\frac{1}{\sqrt{v_{k-1,j}}}-\frac{1}{\sqrt{v_{k,j}}}\right].
\end{align*}
For $k=1$, the last term is $\Ex_1[\tvarphi_{0,j}-\tvarphi_{1,j}]$. For $k\ge 2$,
\[
(\partial_jF(x_k))^2
\Ex_k\left[\frac{1}{\sqrt{v_{k-1,j}}}-\frac{1}{\sqrt{v_{k,j}}}\right]
=
\Ex_k[\tvarphi_{k-1,j}-\tvarphi_{k,j}]
+
\frac{(\partial_jF(x_k))^2-(\partial_jF(x_{k-1}))^2}{\sqrt{v_{k-1,j}}}.
\]
Moreover,
\[
|\partial_jF(x_k)-\partial_jF(x_{k-1})|
\le \|\nabla F(x_k)-\nabla F(x_{k-1})\|
\le L\|x_k-x_{k-1}\|
= \alpha L \sqrt{H_{k-1}},
\]
so
\[
(\partial_jF(x_k))^2-(\partial_jF(x_{k-1}))^2
\le L^2\alpha^2 H_{k-1} + 2\alpha L |\partial_jF(x_k)|\sqrt{H_{k-1}}.
\]
Applying Young's inequality with parameter $\rho$ yields
\[
\tau \alpha L \frac{|\partial_jF(x_k)|}{\sqrt{v_{k-1,j}}}\sqrt{H_{k-1}}
\le
\frac{1}{2\rho}\frac{(\partial_jF(x_k))^2}{\sqrt{v_{k-1,j}}}
+
\frac{\rho\tau^2L^2\alpha^2}{2}\frac{H_{k-1}}{\sqrt{v_{k-1,j}}}.
\]
Therefore, for every $k\in\setK$ and $j\in\setd$,
\begin{align*}
T_{k,j}
\le
\frac12\left(1+\frac1\rho\right)\frac{(\partial_jF(x_k))^2}{\sqrt{v_{k-1,j}}}
+
\frac{\tau}{2}\Ex_k[\tvarphi_{k-1,j}-\tvarphi_{k,j}] 
+
\frac{\tau L^2\alpha^2}{2}(1+\rho\tau)\frac{H_{k-1}}{\sqrt{v_{k-1,j}}}.
\end{align*}

Substituting this into the descent inequality, taking full expectation, and summing over $k=1,\ldots,K$, we obtain
\begin{multline*}
\frac{\alpha}{2}\left(1-\frac1\rho\right)
\sum_{k=1}^K \sum_{j=1}^d
\Ex\left[\frac{(\partial_jF(x_k))^2}{\sqrt{v_{k-1,j}}}\right] \\
\le
F(x_1)-F^\star
+
\frac{\tau\alpha}{2}\sum_{j=1}^d \tvarphi_{0,j}
+
\frac{\tau L^2\alpha^3}{2}(1+\rho\tau)
\sum_{k=1}^K \Ex[\Gamma_{k-1}H_{k-1}]
+
\frac{L\alpha^2}{2}\sum_{k=1}^K \Ex[H_k].
\end{multline*}
Since $v_{k,j}$ is nondecreasing in $k$, $\Gamma_{k-1}\le \Gamma_0$ for all $k$.

By \Cref{lem:seq}, for each $j\in\setd$,
\[
\sum_{k=1}^K \frac{g_{k,j}^2}{v_{k,j}}
\le \log v_{K,j}-\log v_{0,j}.
\]
Hence
\[
\sum_{k=1}^K \Ex[H_k]
=
\sum_{j=1}^d \sum_{k=1}^K \Ex\left[\frac{g_{k,j}^2}{v_{k,j}}\right]
\le
\sum_{j=1}^d \Ex[\log v_{K,j}] - \sum_{j=1}^d \log v_{0,j},
\]
and
\[
\sum_{k=1}^K \Ex[\Gamma_{k-1}H_{k-1}]
\le
\Gamma_0 \sum_{k=1}^{K-1}\Ex[H_k]
\le
\Gamma_0 \sum_{k=1}^K \Ex[H_k].
\]
Therefore there exist constants $c_1,c_2>0$, independent of $K$ and $\delta$, such that
\[
S_K
\coloneqq
\sum_{k=1}^K \sum_{j=1}^d
\Ex\left[\frac{(\partial_jF(x_k))^2}{\sqrt{v_{k-1,j}}}\right]
\le
c_1 + c_2 \sum_{j=1}^d \Ex[\log v_{K,j}].
\]

Now set
\[
Y_K \coloneqq \sum_{j=1}^d \sqrt{v_{K,j}},
\qquad
y_K \coloneqq \Ex[Y_K].
\]
Using the coordinate-wise second-moment bound,
\[
S_K
\ge
\frac1\tau
\sum_{k=1}^K \sum_{j=1}^d
\Ex\left[\frac{g_{k,j}^2}{\sqrt{v_{k-1,j}}}\right].
\]
For each fixed $j$,
\[
\frac{g_{k,j}^2}{\sqrt{v_{k-1,j}}}
=
\frac{v_{k,j}-v_{k-1,j}}{\sqrt{v_{k-1,j}}}
\ge
2(\sqrt{v_{k,j}}-\sqrt{v_{k-1,j}}),
\]
and hence
\[
S_K
\ge
\frac{2}{\tau}\left(y_K-\sum_{j=1}^d \sqrt{v_{0,j}}\right).
\]
On the other hand, pointwise,
\[
\sum_{j=1}^d \log v_{K,j}
\le
2d \log(1+Y_K).
\]
Taking expectation and using Jensen's inequality gives
\[
\sum_{j=1}^d \Ex[\log v_{K,j}]
\le
2d \log(1+y_K).
\]
Combining the last three displays yields
\[
y_K \le c_3 + c_4 \log(1+y_K)
\]
for some constants $c_3,c_4>0$ independent of $K$ and $\delta$. Applying \Cref{lem:thm_AdaGrad-Norm_2} to $1+y_K$ shows that $y_K\le c_5$ for some constant $c_5>0$ independent of $K$ and $\delta$. Consequently,
\[
S_K \le c_6
\]
for some constant $c_6>0$ independent of $K$ and $\delta$.

Finally, since $v_{k-1,j}\le v_{K,j}$, we have pointwise
\[
\sum_{j=1}^d \frac{(\partial_jF(x_k))^2}{\sqrt{v_{k-1,j}}}
\ge
\frac{\|\nabla F(x_k)\|^2}{Y_K}.
\]
Therefore,
\[
S_K
\ge
\Ex\left[\frac{1}{Y_K}\sum_{k=1}^K \|\nabla F(x_k)\|^2\right].
\]
By Cauchy-Schwarz,
\[
\Ex\left[\sqrt{\sum_{k=1}^K \|\nabla F(x_k)\|^2}\right]^2
\le
\Ex[Y_K]\,
\Ex\left[\frac{1}{Y_K}\sum_{k=1}^K \|\nabla F(x_k)\|^2\right]
\le
y_K S_K
\le
c_5c_6.
\]
Set $C\coloneqq c_5c_6$. Then, by Markov's inequality,
\[
\Prob\left(\min_{k\in\setK}\|\nabla F(x_k)\|^2 > \frac{C}{K\delta^2}\right)
\le
\Prob\left(\sqrt{\sum_{k=1}^K \|\nabla F(x_k)\|^2} > \frac{\sqrt{C}}{\delta}\right)
\le
\delta.
\]
Thus, with probability at least $1-\delta$,
\[
\min_{k\in\setK}\|\nabla F(x_k)\|^2 \le \frac{C}{K\delta^2},
\]
which proves the claim.

\subsubsection{Proof of \Cref{thm:AdaGrad-Norm_gen_smooth}}

In case (i), conditional unbiasedness is assumed. In case (ii), $g_k$ is the average of conditionally i.i.d.~per-sample gradients with conditional mean $\nabla F(x_k)$, so it is conditionally unbiased as well. Hence, in both cases,
\[
    \Ex_k[g_k]=\nabla F(x_k),
    \qquad
    \Ex_k[\|g_k\|^2] \le \tau \|\nabla F(x_k)\|^2,
    \qquad k\in\setK,
\]
where the second inequality follows from \Cref{prop:norm,prop:inner_prod}.

Set $x_0 \coloneqq x_1$, $g_0 \coloneqq 0$, $\Delta_k \coloneqq x_k - x_{k-1}$ for $k\in\NN^*$, and
\[
    (\forall k\in\NN)\qquad
    \varphi_k \coloneqq \frac{\|\nabla F(x_k)\|^2}{\sqrt{v_k}}.
\]
Since $1/\rho_1+\rho_1/\rho_2+2\omega<1$, we have $\rho_1>1$ and $\omega<1/2$. As $\tau\ge 1$, the step-size assumption implies $\alpha\le 1/L_1$. Moreover,
\[
    \|x_{k+1}-x_k\|
    = \alpha \left\|\frac{g_k}{\sqrt{v_k}}\right\|
    \le \alpha
    \le \frac{1}{L_1},
\]
because $v_k=v_{k-1}+\|g_k\|^2\ge \|g_k\|^2$. Hence \Cref{lem:thm_AdaGrad-Norm_gen_smooth} gives, pathwise,
\[
    F(\xkpo)
    \le F(x_k) - \alpha \lrdotp{\gradFxk}{\frac{g_k}{\sqrt{v_k}}}
    + \frac{\alpha^2}{2}(L_0 + L_1\|\gradFxk\|)\left\|\frac{g_k}{\sqrt{v_k}}\right\|^2.
\]
Taking conditional expectation yields
\begin{equation}\label{eq:gen_smooth_descent}
    \Ex_k[F(\xkpo)]
    \le F(x_k) - \alpha \lrdotp{\gradFxk}{\Ex_k\left[\frac{g_k}{\sqrt{v_k}}\right]}
    + \frac{L_0\alpha^2}{2}\Ex_k\left[\frac{\|g_k\|^2}{v_k}\right]
    + \frac{L_1\alpha^2\|\gradFxk\|}{2}\Ex_k\left[\frac{\|g_k\|^2}{v_k}\right].
\end{equation}

Using
\[
    \lrdotp{\gradFxk}{\Ex_k\left[\frac{g_k}{\sqrt{v_k}}\right]}
    = \frac{\|\gradFxk\|^2}{\sqrt{\vkmo}}
    + \lrdotp{\gradFxk}{\Ex_k\left[\left(\frac{1}{\sqrt{v_k}}-\frac{1}{\sqrt{\vkmo}}\right)g_k\right]},
\]
we obtain
\[
    -\alpha \lrdotp{\gradFxk}{\Ex_k\left[\frac{g_k}{\sqrt{v_k}}\right]}
    = -\alpha \frac{\|\gradFxk\|^2}{\sqrt{\vkmo}}
    + \alpha \lrdotp{\gradFxk}{\Ex_k\left[\left(\frac{1}{\sqrt{\vkmo}}-\frac{1}{\sqrt{v_k}}\right)g_k\right]}.
\]
Define
\[
    Q_k \coloneqq \frac{\|\gradFxk\|}{\sqrt{\vkmo}}
    \Ex_k\left[\frac{\|g_k\|^2}{\sqrt{v_k}+\sqrt{\vkmo}}\right].
\]
By the identity
\[
    \frac{1}{\sqrt{\vkmo}}-\frac{1}{\sqrt{v_k}}
    = \frac{\|g_k\|^2}{\sqrt{\vkmo}\sqrt{v_k}(\sqrt{v_k}+\sqrt{\vkmo})},
\]
together with $v_k\ge \|g_k\|^2$, we have
\[
    \lrdotp{\gradFxk}{\Ex_k\left[\left(\frac{1}{\sqrt{\vkmo}}-\frac{1}{\sqrt{v_k}}\right)g_k\right]}
    \le Q_k.
\]
Also, since $v_k\ge \vkmo$,
\[
    \frac{1}{2v_k}\le \frac{1}{\sqrt{\vkmo}(\sqrt{v_k}+\sqrt{\vkmo})},
\]
and therefore
\[
    \frac{L_1\alpha^2\|\gradFxk\|}{2}\Ex_k\left[\frac{\|g_k\|^2}{v_k}\right]
    \le L_1\alpha^2 Q_k.
\]
Since $\alpha\le 1/L_1$, \eqref{eq:gen_smooth_descent} becomes
\begin{equation}\label{eq:gen_smooth_descent_2}
    \Ex_k[F(\xkpo)]
    \le F(x_k) - \alpha \frac{\|\gradFxk\|^2}{\sqrt{\vkmo}}
    + 2\alpha Q_k
    + \frac{L_0\alpha^2}{2}\Ex_k\left[\frac{\|g_k\|^2}{v_k}\right].
\end{equation}

Next, by Young's inequality, Cauchy--Schwarz, the E-SG bound, and the inequality
\[
    \frac{\|g_k\|^2}{\sqrt{\vkmo}(\sqrt{v_k}+\sqrt{\vkmo})^2}
    \le \frac{1}{\sqrt{\vkmo}}-\frac{1}{\sqrt{v_k}},
\]
we get
\begin{align*}
    2\alpha Q_k
    &\le \frac{\alpha}{\rho_1}\frac{\|\gradFxk\|^2}{\sqrt{\vkmo}}
    + \frac{\rho_1\alpha}{\sqrt{\vkmo}}
    \left(\Ex_k\left[\frac{\|g_k\|^2}{\sqrt{v_k}+\sqrt{\vkmo}}\right]\right)^2 \\
    &\le \frac{\alpha}{\rho_1}\frac{\|\gradFxk\|^2}{\sqrt{\vkmo}}
    + \frac{\rho_1\alpha}{\sqrt{\vkmo}}
    \Ex_k[\|g_k\|^2]\,
    \Ex_k\left[\frac{\|g_k\|^2}{(\sqrt{v_k}+\sqrt{\vkmo})^2}\right] \\
    &\le \frac{\alpha}{\rho_1}\frac{\|\gradFxk\|^2}{\sqrt{\vkmo}}
    + \rho_1\alpha\tau\|\gradFxk\|^2
    \Ex_k\left[\frac{1}{\sqrt{\vkmo}}-\frac{1}{\sqrt{v_k}}\right].
\end{align*}
We decompose the last term as
\[
    \|\gradFxk\|^2\Ex_k\left[\frac{1}{\sqrt{\vkmo}}-\frac{1}{\sqrt{v_k}}\right]
    = \Ex_k[\varphi_{k-1}-\varphi_k]
    + \frac{\|\gradFxk\|^2-\|\nabla F(\xkmo)\|^2}{\sqrt{\vkmo}}.
\]
By \Cref{ass:gen_smooth},
\[
    \|\gradFxk-\nabla F(\xkmo)\|
    \le (L_0+L_1\|\gradFxk\|)\|\Delta_k\|,
\]
hence, by the triangle inequality and the reverse triangle inequality,
\[
    \|\gradFxk\|^2-\|\nabla F(\xkmo)\|^2
    \le (L_0+L_1\|\gradFxk\|)^2\|\Delta_k\|^2
    + 2(L_0+L_1\|\gradFxk\|)\|\gradFxk\|\|\Delta_k\|.
\]
Using $(a+b)^2\le 2(a^2+b^2)$, we obtain
\begin{align*}
    &\rho_1\alpha\tau
    \frac{\|\gradFxk\|^2-\|\nabla F(\xkmo)\|^2}{\sqrt{\vkmo}} \\
    &\le \rho_1\alpha\tau
    \frac{2L_0^2\|\Delta_k\|^2 + 2L_1^2\|\gradFxk\|^2\|\Delta_k\|^2 + 2L_0\|\gradFxk\|\|\Delta_k\| + 2L_1\|\gradFxk\|^2\|\Delta_k\|}{\sqrt{\vkmo}}.
\end{align*}
Using
\[
    2\tau L_0\|\gradFxk\|\|\Delta_k\|
    \le \frac{1}{\rho_2}\|\gradFxk\|^2 + \rho_2\tau^2 L_0^2\|\Delta_k\|^2,
\]
together with $\|\Delta_k\|\le \alpha$ and the step-size restriction
\[
    2\rho_1\tau L_1\|\Delta_k\|\le \omega,
    \qquad
    2\rho_1\tau L_1^2\|\Delta_k\|^2\le \omega,
\]
we obtain
\[
    \rho_1\alpha\tau
    \frac{\|\gradFxk\|^2-\|\nabla F(\xkmo)\|^2}{\sqrt{\vkmo}}
    \le \alpha\left(\frac{\rho_1}{\rho_2}+2\omega\right)\frac{\|\gradFxk\|^2}{\sqrt{\vkmo}}
    + \rho_1L_0^2\tau(2+\rho_2\tau)\alpha\frac{\|\Delta_k\|^2}{\sqrt{\vkmo}}.
\]
Since $\Delta_k=-\alpha \gkmo/\sqrt{\vkmo}$, this yields
\begin{multline}\label{eq:gen_smooth_Qk}
    2\alpha Q_k
    \le \alpha\left(\frac{1}{\rho_1}+\frac{\rho_1}{\rho_2}+2\omega\right)\frac{\|\gradFxk\|^2}{\sqrt{\vkmo}}
    + \rho_1\alpha\tau\Ex_k[\varphi_{k-1}-\varphi_k] \\
    + \rho_1L_0^2\tau(2+\rho_2\tau)\alpha^3\frac{\|\gkmo\|^2}{\vkmo^{3/2}}.
\end{multline}
Combining \eqref{eq:gen_smooth_descent_2} and \eqref{eq:gen_smooth_Qk}, and setting
\[
    c \coloneqq 1-\frac{1}{\rho_1}-\frac{\rho_1}{\rho_2}-2\omega > 0,
\]
we obtain
\begin{multline}\label{eq:gen_smooth_final_descent}
    \Ex_k[F(\xkpo)]
    \le F(x_k) - c\alpha\frac{\|\gradFxk\|^2}{\sqrt{\vkmo}}
    + \rho_1\alpha\tau\Ex_k[\varphi_{k-1}-\varphi_k] \\
    + \frac{L_0\alpha^2}{2}\Ex_k\left[\frac{\|g_k\|^2}{v_k}\right]
    + \rho_1L_0^2\tau(2+\rho_2\tau)\alpha^3\frac{\|\gkmo\|^2}{\vkmo^{3/2}}.
\end{multline}

Taking total expectation and summing \eqref{eq:gen_smooth_final_descent} over $k=1,\ldots,K$, then using $F(x_{K+1})\ge F^\star$, $\varphi_K\ge 0$, and \Cref{lem:seq}, gives
\[
    \sum_{k=1}^K \Ex\left[\frac{\|\gradFxk\|^2}{\sqrt{\vkmo}}\right]
    \le A + B\,\Ex[\log v_K]
\]
for some finite constants $A,B>0$ independent of $K$ and $\delta$. Indeed,
\[
    \sum_{k=1}^K \frac{\|g_k\|^2}{v_k}\le \log v_K-\log v_0,
    \qquad
    \sum_{k=1}^K \frac{\|\gkmo\|^2}{\vkmo^{3/2}}\le \frac{2}{\sqrt{v_0}},
\]
and $\sum_{k=1}^K \Ex[\varphi_{k-1}-\varphi_k]=\Ex[\varphi_0-\varphi_K]\le \varphi_0$.

Next,
\begin{align*}
    \Ex[\sqrt{v_K}]
    &= \sqrt{v_0}
    + \sum_{k=1}^K \Ex\left[\frac{\|g_k\|^2}{\sqrt{v_k}+\sqrt{\vkmo}}\right] \\
    &\le \sqrt{v_0}
    + \frac12\sum_{k=1}^K \Ex\left[\frac{\|g_k\|^2}{\sqrt{\vkmo}}\right] \\
    &\le \sqrt{v_0}
    + \frac{\tau}{2}\sum_{k=1}^K \Ex\left[\frac{\|\gradFxk\|^2}{\sqrt{\vkmo}}\right] \\
    &\le \sqrt{v_0} + \frac{\tau}{2}\left(A+B\,\Ex[\log v_K]\right).
\end{align*}
Since $\Ex[\log v_K]=2\Ex[\log \sqrt{v_K}]\le 2\log \Ex[\sqrt{v_K}]$ by Jensen's inequality, \Cref{lem:thm_AdaGrad-Norm_2} implies that $\Ex[\sqrt{v_K}]\le C_v$ for some finite constant $C_v>0$ independent of $K$. Consequently,
\[
    \sum_{k=1}^K \Ex\left[\frac{\|\gradFxk\|^2}{\sqrt{\vkmo}}\right]
    \le A+2B\log C_v \eqqcolon C_s.
\]

Since $v_{k-1}\le v_K$, we have
\[
    \sum_{k=1}^K \Ex\left[\frac{\|\gradFxk\|^2}{\sqrt{\vkmo}}\right]
    \ge \Ex\left[\frac{1}{\sqrt{v_K}}\sum_{k=1}^K \|\gradFxk\|^2\right].
\]
By Cauchy--Schwarz,
\[
    \Ex\left[\sqrt{\sum_{k=1}^K \|\gradFxk\|^2}\right]^2
    \le \Ex[\sqrt{v_K}]
    \Ex\left[\frac{1}{\sqrt{v_K}}\sum_{k=1}^K \|\gradFxk\|^2\right]
    \le C_v C_s.
\]
Finally, since
\[
    \sum_{k=1}^K \|\gradFxk\|^2 \ge K\min_{k\in\setK}\|\nabla F(x_k)\|^2,
\]
Markov's inequality yields
\[
    \Prob\left(
        \min_{k\in\setK}\|\nabla F(x_k)\|^2
        > \frac{C_v C_s}{K\delta^2}
    \right)
    \le \Prob\left(
        \sqrt{\sum_{k=1}^K \|\gradFxk\|^2}
        > \frac{\sqrt{C_v C_s}}{\delta}
    \right)
    \le \delta.
\]
Therefore, with probability at least $1-\delta$,
\[
    \min_{k\in\setK}\|\nabla F(x_k)\|^2
    \le \frac{C_v C_s}{K\delta^2}.
\]
This proves the claim with $C\coloneqq C_v C_s$.
    
\section{Details and Additional Results of Numerical Experiments}
\label{sec:expt_details}

In this section, we provide details and additional results for the numerical experiments in \Cref{sec:expt}. In particular, we report the training hyperparameters used in the experiments and include additional plots not shown in the main text.

\paragraph{Sampling protocol, test frequency, and numerical safeguards.}
All experiments use the standard shuffled-epoch sampling protocol. At the beginning of each epoch, the training data are randomly shuffled, and minibatches are then processed sequentially from the shuffled data order. Thus, the empirical implementation follows the usual without-replacement shuffled-epoch protocol rather than independent sampling with replacement. In the adaptive-batch experiments, the batch-size statistic computed from the current minibatch is used to determine the batch size for the next adaptive update. The adaptive test is evaluated once per optimizer iteration during the adaptive phase, except for the three-layer CNN on CIFAR-10 experiment, where it is evaluated every $10$ optimizer iterations. Once the prescribed maximum batch size is reached, we stop evaluating the adaptive test and continue training with the capped batch size. The initial (base) batch sizes and maximum batch sizes for all experiments are reported in their corresponding training hyperparameter tables. We did not apply denominator clipping or any other special numerical treatment to potentially small empirical gradient norms appearing in the denominators of the batch-size test statistics. Empirically, the maximum batch size was typically reached at an early or intermediate stage of training, so near-stationary regimes with very small empirical gradient norms did not require denominator clipping or regularization in our reported experiments.

\paragraph{Aggregate cost reporting.}
To make the computational comparison more transparent, we report aggregate cost statistics for the experimental runs in individual subsections. The reported wall-clock time is the end-to-end training time and includes the cost of computing the empirical adaptive-test statistics, including the per-sample gradient quantities used by the tests. However, our implementation does not separately instrument the forward/backward pass, the per-sample-gradient-statistic computation, the batch-size-test computation, or GPU memory usage. Thus, the reported timings should be interpreted as aggregate cost measurements rather than as an isolated decomposition of adaptive-test overhead or memory overhead.

\subsection{Multi-class Logistic Regression on MNIST}    

\Cref{table:hyperparams_log_reg} lists the training specifications and optimizer hyperparameters for the experiments on multi-class logistic regression on the MNIST dataset.
    
\begin{table}[h!]
    \centering
    \caption{Training hyperparameters for multi-class logistic regression on MNIST}
    \label{table:hyperparams_log_reg}
    \begin{tabular}{lc}
        Model & Multi-class Logistic Regression \\
        \midrule
        Training budget & 6,000,000 samples (100 epochs) \\
        Weight initialization & Default \\
        Learning rate schedule & None \\
        Optimizer & \SGD or \textsc{AdaGrad}(-\textsc{Norm}) \\
        Base learning rate & 0.008 \\
        Base batch size & 2 \\
        Maximum global batch size & 60,000 \\
        Weight decay & 0 \\
        Momentum & 0 \\
        Precision & \texttt{tf32} \\
        \bottomrule
    \end{tabular}
\end{table}

\begin{figure}[h!]
        \centering
        \includegraphics[width=0.7\textwidth]{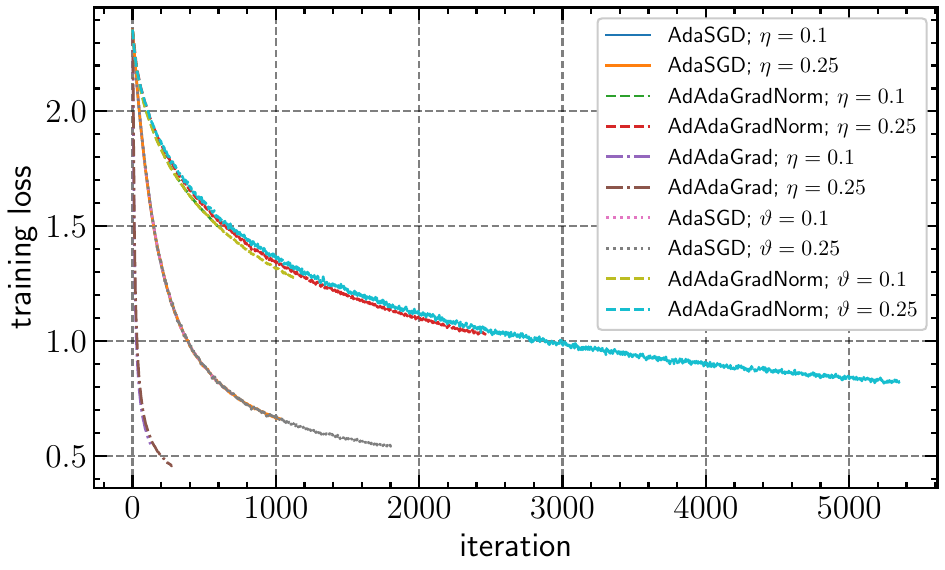}
        \caption{Training loss curves (vs.~number of iterations) of \AdaSGD, \AdAdaGrad, and \AdAdaGradNorm for logistic regression on the MNIST dataset.}
        \label{fig:log_reg_mnist_iteration}
    \end{figure}

\Cref{table:log_reg_cost} provides aggregate cost statistics for the experiments on multi-class logistic regression on the MNIST dataset, and we also plot the training loss curves against the number of iterations in \Cref{fig:log_reg_mnist_iteration}.

\begin{table}[h!]
        \centering
        \caption{Aggregate cost statistics for multi-class logistic regression on MNIST}
        \label{table:log_reg_cost}
        \small
        \begin{tabular}{ccrrrrrr}        
            \toprule
            Scheme & test & time (h) & samples & steps & batch size & loss & accuracy \\
            \midrule
            \AdaSGD & $\eta=0.10$ & 0.40 & 6,030,188 & 351 & 17131 &  1.04 &  0.82  \\
            \AdaSGD & $\eta=0.25$  & 0.91 & 6,006,252 & 1029  & 5831 &  0.67 &  0.86 \\
            \AdAdaGradNorm & $\eta=0.10$ & 0.63 & 6,005,933 & 596 & 10060 & 1.50  &  0.78  \\
            \AdAdaGradNorm & $\eta=0.25$ & 1.91 & 6,001,106 & 2462 & 2437 & 1.02 &  0.83  \\
            \AdAdaGrad & $\eta=0.10$  & \textbf{0.15} & \textbf{6,004,856} & \textbf{126} & \textbf{47282} & \textbf{0.54} & \textbf{0.88} \\
            \AdAdaGrad & $\eta=0.25$  & \textbf{0.30} & \textbf{6,027,458} & \textbf{274} & \textbf{21918} & \textbf{0.46} & \textbf{0.90} \\
            \AdaSGD & $\vartheta=0.01$  & 1.11 & 6,003,750 & 717 & 8362 & 0.76 & 0.85   \\
            \AdaSGD & $\vartheta=0.05$  & 2.13 & 6,005,454 & 1804 & 3327 & 0.54 & 0.88   \\
            \AdAdaGradNorm & $\vartheta=0.01$ & 1.50 & 6,002,839 & 1127 & 5322 & 1.28 &   0.80 \\
            \AdAdaGradNorm & $\vartheta=0.05$ & 5.37 & 6,001,053 & 5349 & 1122 & 0.80  &  0.85  \\
            \bottomrule
        \end{tabular}    
    \end{table} 	

\newpage
\subsection{Three-layer Convolutional Neural Network on MNIST}
\label{subsec:cnn_mnist_supp}
\Cref{table:hyperparams_cnn_mnist} lists the training specifications and optimizer hyperparameters for the experiments on a three-layer convolutional neural network on the MNIST dataset.

\begin{table}[h!]
    \centering
    \caption{Training hyperparameters for three-layer CNN on MNIST}
    \label{table:hyperparams_cnn_mnist}
    \begin{tabular}{lc}
        Model & 3-layer CNN on MNIST \\
        \midrule
        Training budget & 6,000,000 samples (100 epochs) \\
        Weight initialization & Default \\
        Optimizer & \SGD or \textsc{AdaGrad}(-\textsc{Norm}) \\
        Base learning rate & 0.008 \\
        Base batch size & 8 \\
        Maximum batch size & 60,000 \\
        Weight decay & 0 \\
        Momentum & 0 \\
        Precision & \texttt{tf32} \\
        \bottomrule
    \end{tabular}
\end{table}

\Cref{table:cnn_mnist_full_cost} provides aggregate cost statistics for the experiments on a three-layer convolutional neural network on the MNIST dataset. 

\begin{table}[h!]
    \centering
    \caption{Aggregate cost statistics for three-layer CNN on MNIST}
    \label{table:cnn_mnist_full_cost}
    \small
    \begin{tabular}{ccrrrrrr}
        \toprule
        Scheme & test & time (h) & samples & steps & batch size & loss & accuracy \\
        \midrule
        \SGD & N/A & 6.77 & 6,000,640 & 2929 & 2048 & 0.12 & 0.96 \\
        \SGD & N/A & 3.38 & 6,000,640 & 1464 & 4096 & 0.20 & 0.94 \\
        \SGD & N/A & 1.72 & 6,004,736 & 732 & 8192 & 0.32 & 0.91 \\
        \SGD & N/A & 0.90 & 6,012,928 & 366 & 16384 & 0.51 & 0.87 \\
        \SGD & N/A & 0.45 & 6,029,312 & 183 & 32768 & 1.54 & 0.75 \\
        \SGD & N/A & 0.27 & 6,000,000 & 99 & 60000 & 2.15 & 0.66 \\
        \AdaGrad & N/A & 7.12 & 6,000,640 & 2929 & 2048 & 0.02 & 0.99 \\
        \AdaGrad & N/A & 3.60 & 6,000,640 & 1464 & 4096 & 0.02 & 0.99 \\
        \AdaGrad & N/A & 1.82 & 6,004,736 & 732 & 8192 & 0.05 & 0.98 \\
        \AdaGrad & N/A & 0.92 & 6,012,928 & 366 & 16384 & 0.07 & 0.98 \\
        \AdaGrad & N/A & 0.52 & 6,000,000 & 199 & 30000 & 0.10 & 0.97 \\
        \AdaGrad & N/A & 0.47 & 6,029,312 & 183 & 32768 & 0.11 & 0.97 \\
        \AdaGrad & N/A & 0.36 & 6,000,000 & 149 & 40000 & 0.13 & 0.96 \\
        \AdaGrad & N/A & 0.29 & 6,000,000 & 99 & 60000 & 0.17 & 0.95 \\
        \AdaSGD & $\eta=0.10$ & 0.73 & 6,051,456 & 256 & 23546 & 0.79 & 0.83 \\
        \AdaSGD & $\eta=0.25$ & 1.05 & 6,000,868 & 383 & 15627 & 0.48 & 0.88 \\
        \AdAdaGradNorm & $\eta=0.10$ & 0.65 & 6,030,808 & 226 & 26567 & 0.88 & 0.83 \\
        \AdAdaGradNorm & $\eta=0.25$ & 1.27 & 6,030,096 & 435 & 13830 & 0.54 & 0.87 \\
        \AdAdaGrad & $\eta=0.10$ & \textbf{0.45} & \textbf{6,008,592} & \textbf{149} & \textbf{40057} & \textbf{0.15} & \textbf{0.96} \\
        \AdAdaGrad & $\eta=0.25$ & \textbf{0.58} & \textbf{6,000,412} & \textbf{198} & \textbf{30152} & \textbf{0.13} & \textbf{0.97} \\
        \AdAdaGrad & $\eta=0.50$ & \textbf{0.62} & \textbf{6,035,096} & \textbf{215} & \textbf{27940} & \textbf{0.11} & \textbf{0.97} \\
        \AdAdaGrad & $\eta=0.75$ & \textbf{0.79} & \textbf{6,046,180} & \textbf{271} & \textbf{22228} & \textbf{0.10} & \textbf{0.97} \\
        \AdaSGD & $\vartheta=0.01$ & 0.63 & 6,024,044 & 230 & 26078 & 0.98 & 0.80 \\
        \AdaSGD & $\vartheta=0.05$ & 1.17 & 6,047,004 & 411 & 14593 & 0.45 & 0.88 \\
        \AdAdaGradNorm & $\vartheta=0.01$ & 0.70 & 6,019,232 & 241 & 24872 & 0.83 & 0.84 \\
        \AdAdaGradNorm & $\vartheta=0.05$ & 1.44 & 6,012,544 & 528 & 11365 & 0.50 & 0.88 \\
        \bottomrule
    \end{tabular}
\end{table}

\begin{figure*}[h!]
    \centering
    \includegraphics[width=0.8\textwidth]{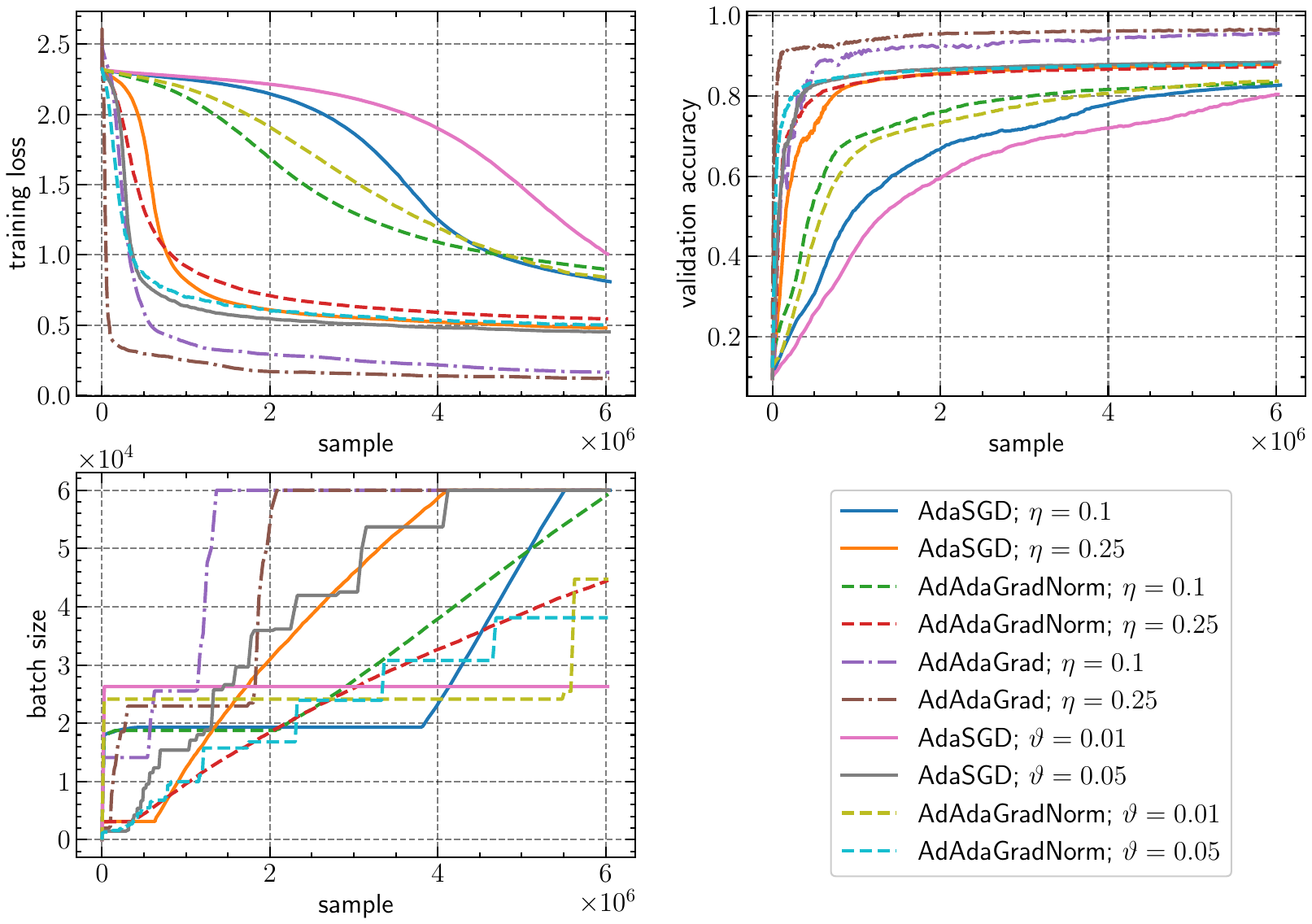}
    \caption{Training loss, validation accuracy and batch size curves (vs.~number of training samples) of \AdaSGD, \AdAdaGrad and \AdAdaGradNorm for three-layer CNN on the MNIST dataset.}
    \label{fig:cnn_mnist}
\end{figure*}

In addition to the training hyperparameters, we provide additional plots for the norm test and the inner product-based test and briefly compare their batch-size dynamics. From \Cref{fig:cnn_mnist}, we observe that, for the runs shown, the norm test yields faster and larger batch-size increases than the inner product test, even for small $\vartheta$. In applications where sharp batch-size increases may induce training instability, the more gradual behavior of the inner product test can be preferable. 

\begin{figure*}[h!]
    \centering
    \includegraphics[width=0.8\textwidth]{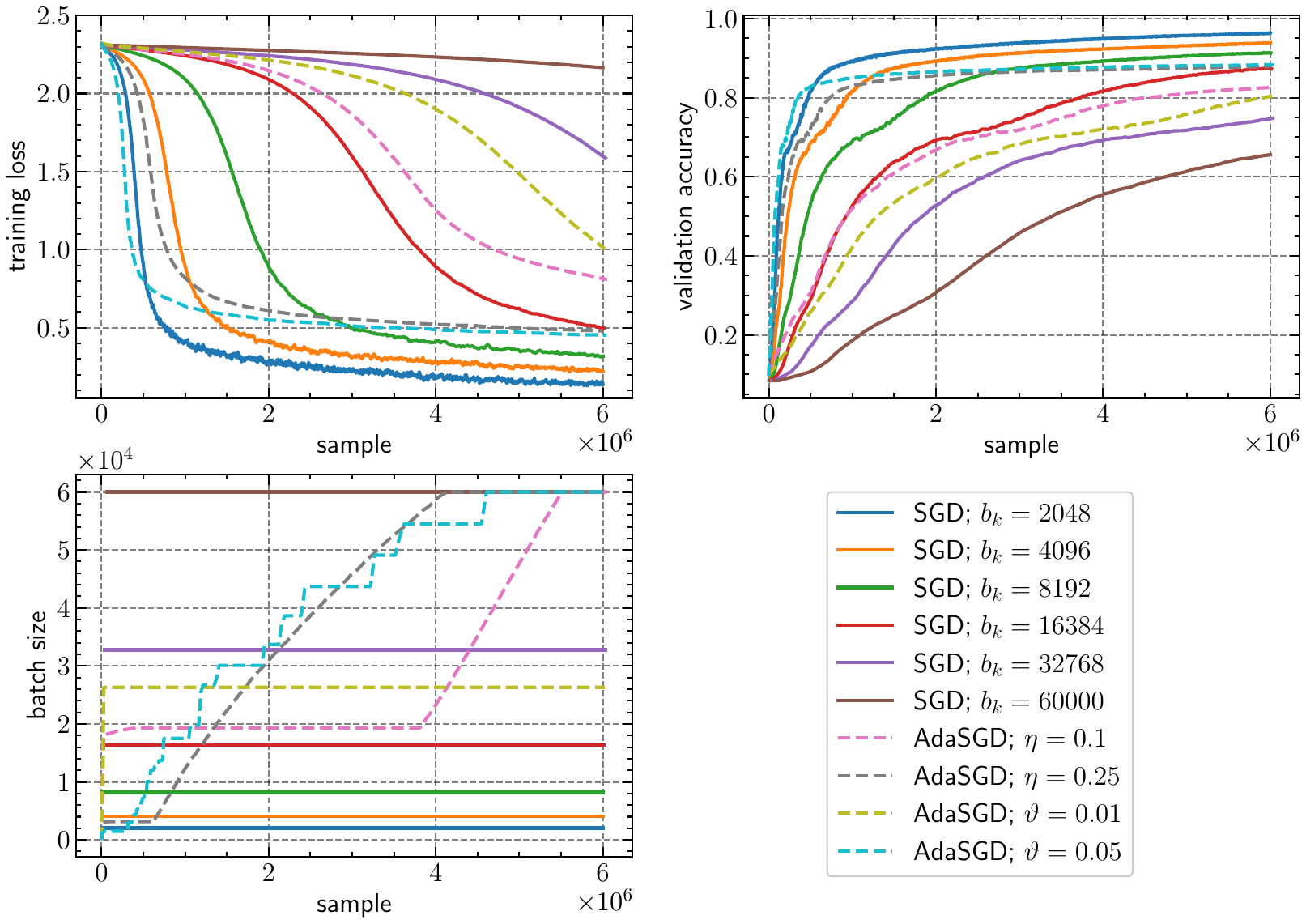}
    \caption{Training loss, validation accuracy and batch size curves (vs.~number of training samples) of \SGD and \AdaSGD for three-layer CNN on the MNIST dataset.}
    \label{fig:cnn_mnist_adasgd}
\end{figure*}

\begin{figure*}[h!]
    \centering
    \includegraphics[width=0.8\textwidth]{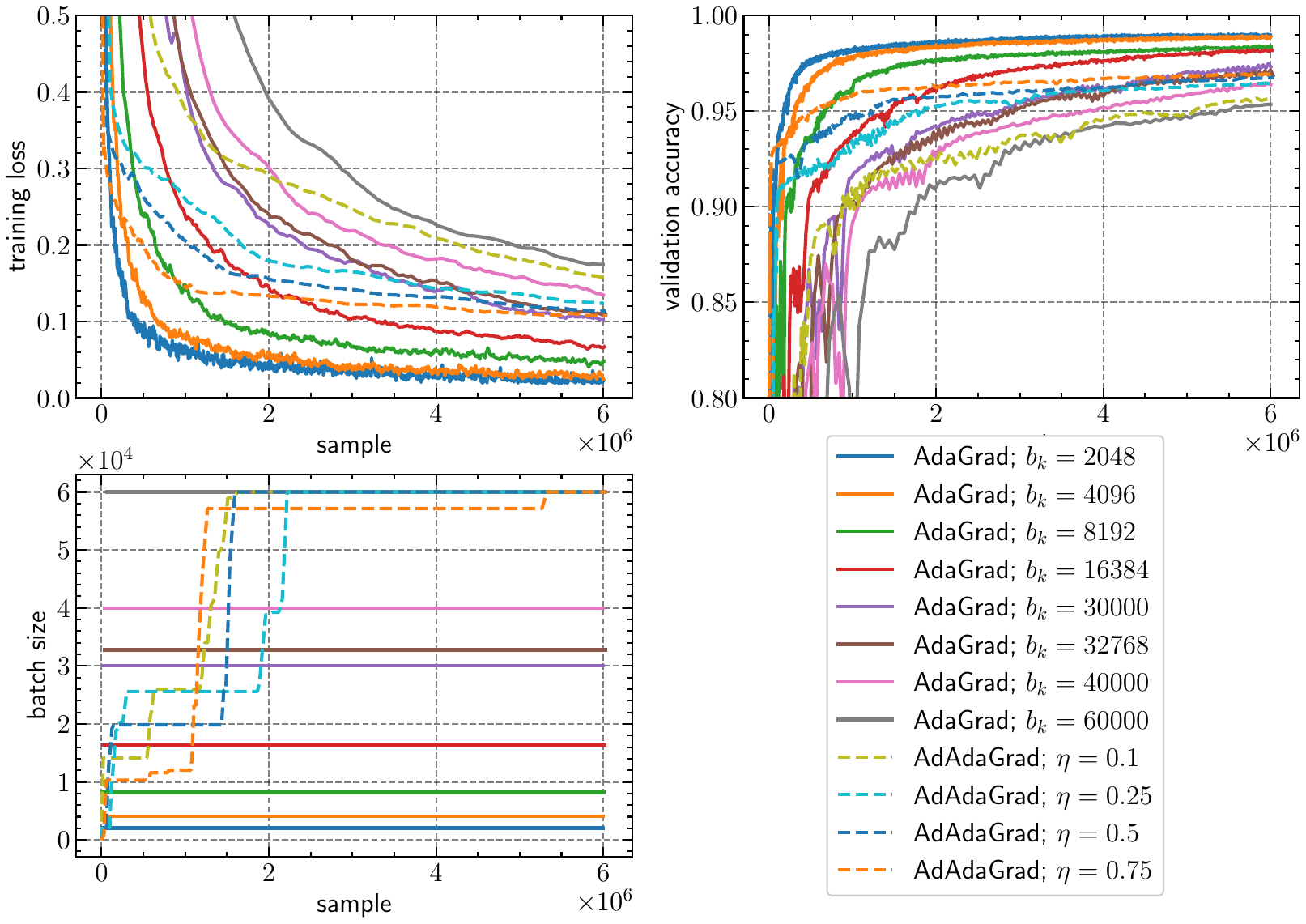}
    \caption{Training loss, validation accuracy and batch size curves (vs.~number of training samples) of \AdaGrad and \AdAdaGrad for three-layer CNN on the MNIST dataset.}
    \label{fig:cnn_mnist_adagrad_2}
\end{figure*}

We also provide two additional sets of plots comparing constant-batch and adaptive-batch variants of \SGD and \AdaGrad, respectively; see \Cref{fig:cnn_mnist_adasgd,fig:cnn_mnist_adagrad_2}. These plots illustrate how adaptive batch sizes based on the norm test or the inner product-based test compare with a range of constant batch sizes. We observe that the adaptive schemes generally attain validation accuracies between those obtained with small and large constant batches, suggesting that they can balance computational efficiency and generalization when larger batches are desired. 

\begin{figure}[h!]
    \centering
    \includegraphics[width=0.7\textwidth]{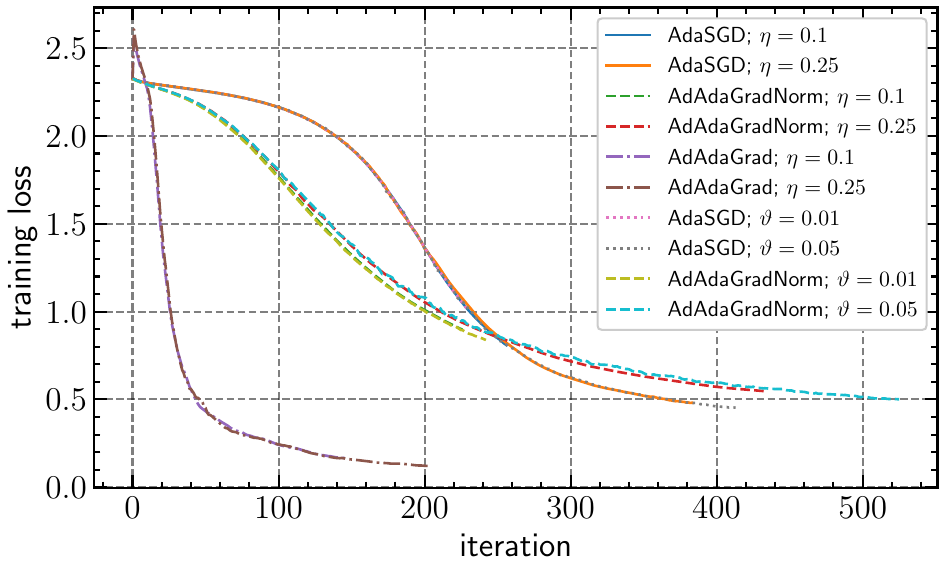}
    \caption{Training loss curves (vs.~number of iterations) of \AdaSGD, \AdAdaGrad, and \AdAdaGradNorm for three-layer CNN on the MNIST dataset.}
    \label{fig:cnn_mnist_iteration}
\end{figure}

\begin{figure*}[h!]
    \centering
    \includegraphics[width=0.7\textwidth]{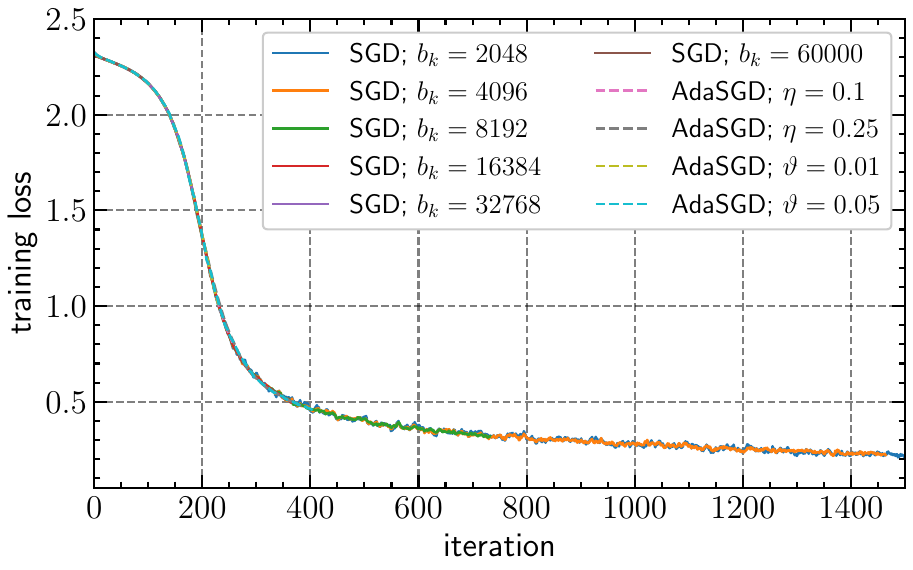}
    \caption{Training loss curves (vs.~number of iterations) of \SGD and \AdaSGD for three-layer CNN on the MNIST dataset.}
    \label{fig:cnn_mnist_adasgd_iteration}
\end{figure*}

\begin{figure*}[h!]
    \centering
    \includegraphics[width=0.7\textwidth]{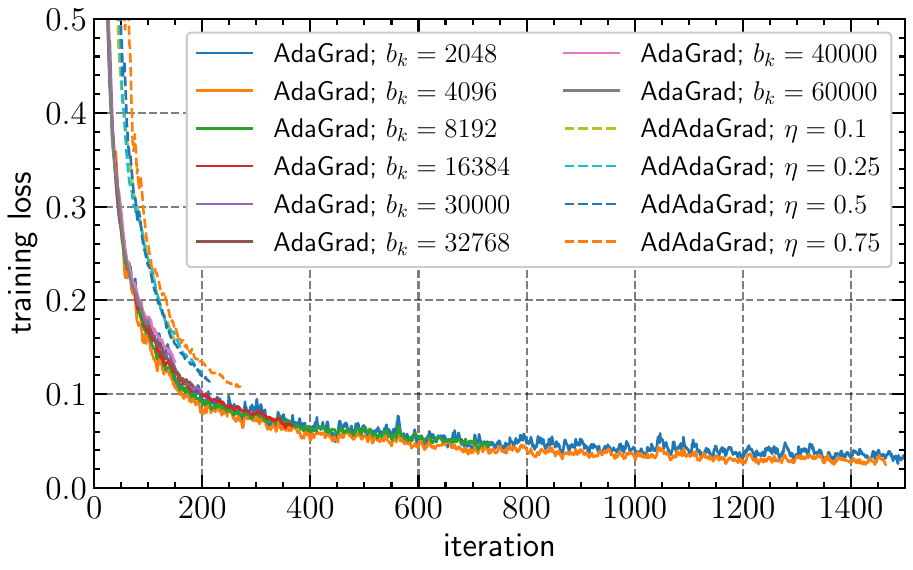}
    \caption{Training loss curves (vs.~number of iterations) of \AdaGrad and \AdAdaGrad for three-layer CNN on the MNIST dataset.}
    \label{fig:cnn_mnist_adagrad_2_iteration}
\end{figure*}

We also give their corresponding iteration-based plots of the training loss curves in \Cref{fig:cnn_mnist_iteration,fig:cnn_mnist_adasgd_iteration,fig:cnn_mnist_adagrad_2_iteration}. 

\clearpage
\newpage
\subsection{Three-layer Convolutional Neural Network on CIFAR-10}
\Cref{table:hyperparams_cnn_cifar-10} lists the training specifications and optimizer hyperparameters for the experiments on a three-layer convolutional neural network on the CIFAR-10 dataset. As noted in the main text, to reduce the computational overhead of adaptive testing, the test is evaluated every 10 steps in these experiments.

\begin{table}[h!]
    \centering
    \caption{Training hyperparameters for three-layer CNN on CIFAR-10}
    \label{table:hyperparams_cnn_cifar-10}
    \begin{tabular}{lc}
        Model & 3-layer CNN on CIFAR-10 \\
        \midrule
        Training budget & 5,000,000 samples (100 epochs) \\
        Weight initialization & Default \\
        Learning rate schedule & None \\
        Optimizer & \SGD or \textsc{AdaGrad}(-\textsc{Norm}) \\
        Optimizer scaling rule & None \\
        Base learning rate & 0.05 \\
        Base batch size & 2 \\
        Maximum batch size & 10,000 \\
        Weight decay & 0 \\
        Momentum & 0 \\
        Precision & \texttt{tf32} \\
        \bottomrule
    \end{tabular}
\end{table}

\Cref{table:cnn_cifar-10_cost} provides aggregate cost statistics for the experiments on a three-layer convolutional neural network on the CIFAR-10 dataset. 

\begin{table}[h!]
    \centering
    \caption{Aggregate cost statistics for three-layer CNN on CIFAR-10}
    \label{table:cnn_cifar-10_cost}
    \small
    \begin{tabular}{ccrrrrrr}
        \toprule
        Scheme & test & time (h) & samples & steps & batch size & loss & accuracy \\
        \midrule
        \AdaSGD & $\eta=0.25$ & 0.87 & 5,001,302 & 523 & 9544 &  1.68 & 0.40  \\
        \AdaSGD & $\eta=0.50$ & 0.96 & 5,003,402 & 658 & 7592 & 1.59  &  0.43  \\
        \AdAdaGradNorm & $\eta=0.25$ & \textbf{0.87} & \textbf{5,001,302} & \textbf{531}  & \textbf{9401} &  \textbf{1.36} &  \textbf{0.52}  \\
        \AdAdaGradNorm & $\eta=0.50$ & \textbf{1.19} & \textbf{5,002,835} & \textbf{1261} & \textbf{3964} & \textbf{1.19}  &  \textbf{0.57}  \\
        \AdAdaGrad & $\eta=0.25$ & 1.05 & 5001581 & 903 & 5533 & 1.20 &  0.54  \\
        \AdAdaGrad & $\eta=0.50$ & \textbf{1.15} & \textbf{5,003,236} & \textbf{1123} & \textbf{4451} & \textbf{1.11}  &  \textbf{0.57}  \\
        \AdaSGD & $\vartheta=0.05$ & \textbf{1.38} & 5,001,178 & \textbf{1597}  & \textbf{3130} & \textbf{1.17}  &  \textbf{0.57}  \\
        \AdaSGD & $\vartheta=0.10$ & 0.96 & 5,003,342 & 640 & 7806 & 1.64  & 0.41  \\
        \AdAdaGradNorm & $\vartheta=0.05$ & \textbf{0.99} & \textbf{5,008,602} & \textbf{780}  & \textbf{6413} &  \textbf{1.29} &  \textbf{0.55}  \\
        \AdAdaGradNorm & $\vartheta=0.10$ & 1.54 & 5,003,894 & 1948  & 2567 &  1.13 & 0.58  \\
        \bottomrule
    \end{tabular}
\end{table} 

\begin{figure}[h!]
    \centering
    \includegraphics[width=0.65\textwidth]{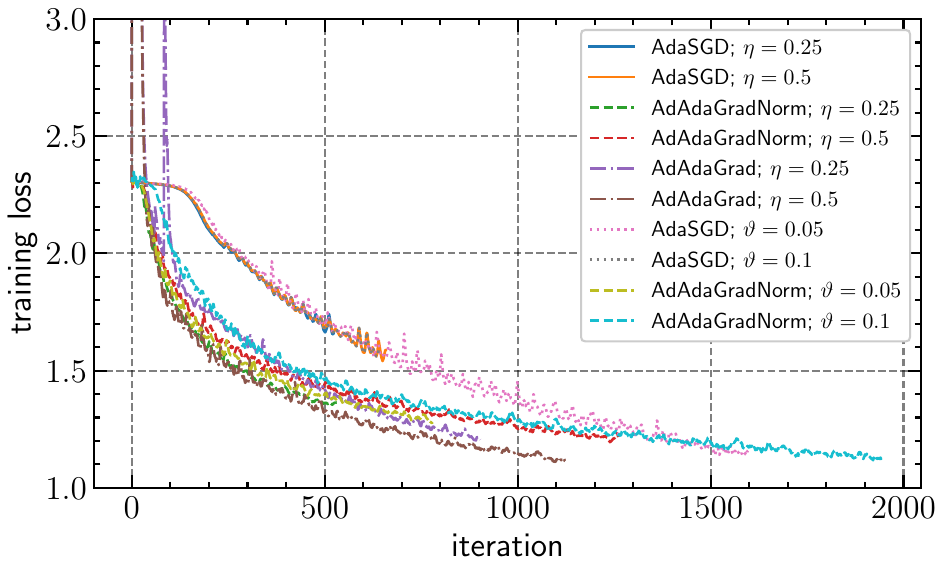}
    \caption{Training loss curves (vs.~number of iterations) of \AdaSGD, \AdAdaGrad, and \AdAdaGradNorm for a three-layer CNN on the CIFAR-10 dataset.}
    \label{fig:cnn_cifar-10_iteration}
\end{figure}

We give the iteration-based plot of the training loss curves in \Cref{fig:cnn_cifar-10_iteration}. 

\newpage
\subsection{\ResNet-18 on CIFAR-10}
\label{subsec:resnet-18_cifar10_supp}
\Cref{table:hyperparams_resnet-18_cifar-10} lists the training specifications and optimizer hyperparameters for the experiments on \ResNet-18 on the CIFAR-10 dataset.

\begin{table}[h!]
    \centering
    \caption{Training hyperparameters for \ResNet-18 on CIFAR-10}
    \label{table:hyperparams_resnet-18_cifar-10}
    \begin{tabular}{lc}
        Model & \ResNet-18 on CIFAR-10 \\
        \midrule
        Training budget & 10,000,000 samples (200 epochs) \\
        Weight initialization & Default \\
        Optimizer & \AdaGrad or \Adam\\
        Learning rate schedule & Linear warmup + cosine decay \\
        Learning rate warmup (samples) & 1M \\
        $(\beta_1, \beta_2)$ & $(0.9, 0.95)$ \\
        $\varepsilon$ & $10^{-8}$ \\
        Peak learning rate & 0.05 \\
        Minimum learning rate & 0.005 \\
        Base batch size & 8 \\
        Maximum batch size & 50,000 \\
        Weight decay & 0 \\
        Precision & \texttt{tf32} \\
        \bottomrule
    \end{tabular}
\end{table}

\begin{figure}[h!]
    \centering
    \includegraphics[width=0.65\textwidth]{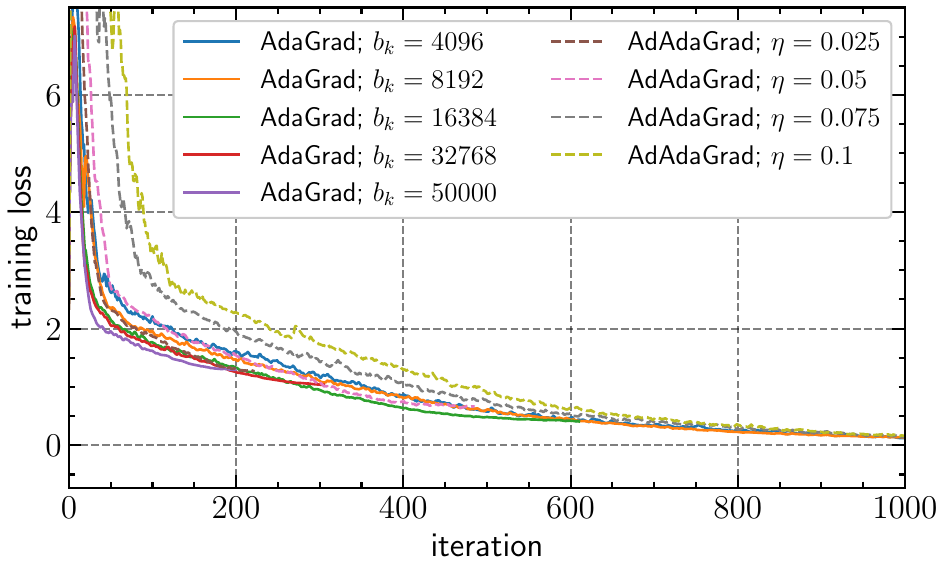}
    \caption{Training loss curves (vs.~number of iterations) of \AdaGrad and \AdAdaGrad for \ResNet-18 on the CIFAR-10 dataset.}
    \label{fig:resnet-18_cifar10_adagrad_iteration}
\end{figure}

\begin{figure}[h!]
    \centering
    \includegraphics[width=0.65\textwidth]{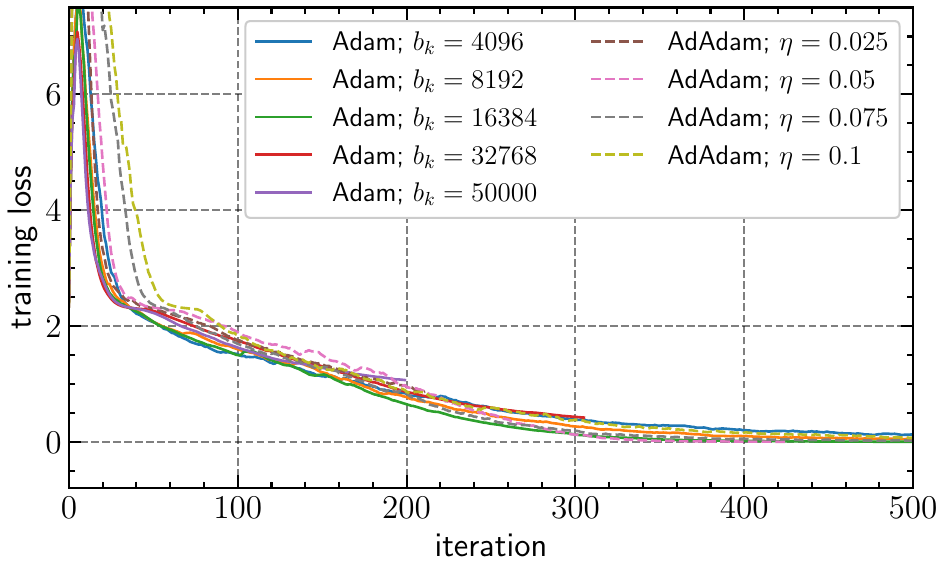}
    \caption{Training loss curves (vs.~number of iterations) of \Adam and \AdAdam for \ResNet-18 on the CIFAR-10 dataset.}
    \label{fig:resnet-18_cifar10_adam_iteration}
\end{figure}

\Cref{table:resnet-18_cifar10_full_cost} provides aggregate cost statistics for the experiments on \ResNet-18 on the CIFAR-10 dataset, while iteration-based plots of the training loss curves are given in \Cref{fig:resnet-18_cifar10_adagrad_iteration,fig:resnet-18_cifar10_adam_iteration}. 

\begin{table}[h!]
    \centering
    \caption{Aggregate cost statistics for \ResNet-18 on CIFAR-10}
    \label{table:resnet-18_cifar10_full_cost}
    \small
    \begin{tabular}{ccrrrrrr}
        \toprule
        Scheme & test & time (h) & samples & steps & batch size & loss & accuracy \\
        \midrule
        \AdaGrad & N/A & 1.21 & 10,002,432 & 2441 & 4096 & 0.0042 & 0.8521 \\
        \AdaGrad & N/A & 0.97 & 10,002,432 & 1220 & 8192 & 0.0808 & 0.8072 \\
        \AdaGrad & N/A & 0.77 & 10,010,624 & 610 & 16384 & 0.5098 & 0.7264 \\
        \AdaGrad & N/A & 0.45 & 10,027,008 & 305 & 32768 & 0.9684 & 0.5816 \\
        \AdaGrad & N/A & 0.33 & 10,000,000 & 199 & 50000 & 1.3625 & 0.4708 \\
        \Adam & N/A & 1.20 & 10,002,432 & 2441 & 4096 & 0.0003 & 0.9147 \\
        \Adam & N/A & 0.97 & 10,002,432 & 1220 & 8192 & 0.0004 & 0.8946 \\
        \Adam & N/A & 0.77 & 10,010,624 & 610 & 16384 & 0.0028 & 0.8628 \\
        \Adam & N/A & 0.45 & 10,027,008 & 305 & 32768 & 0.4000 & 0.7463 \\
        \Adam & N/A & 0.33 & 10,000,000 & 199 & 50000 & 1.0680 & 0.5750 \\
        \AdAdaGrad & $\eta=0.025$ & 0.32 & 10,020,200 & 222 & 44934 & 1.2770 & 0.5107 \\
        \AdAdaGrad & $\eta=0.050$ & 0.60 & 10,018,824 & 485 & 20615 & 0.6204 & 0.7079 \\
        \AdAdaGrad & $\eta=0.075$ & \textbf{1.02} & \textbf{10,004,880} & \textbf{1697} & \textbf{5892} & \textbf{0.0258} & \textbf{0.8180} \\
        \AdAdaGrad & $\eta=0.100$ & \textbf{0.94} & \textbf{10,008,128} & \textbf{1404} & \textbf{7123} & \textbf{0.0668} & \textbf{0.8085} \\
        \AdAdam & $\eta=0.025$ & 0.34 & 10,044,552 & 211 & 47380 & 0.9039 & 0.6234 \\
        \AdAdam & $\eta=0.050$ & 0.60 & 10,018,824& 426 & 23463 & 0.0061 & 0.8228 \\
        \AdAdam & $\eta=0.075$ & 0.74 & 10,008,248 & 900 & 11108 & 0.0008 & 0.8983 \\
        \AdAdam & $\eta=0.100$ & \textbf{0.82} & \textbf{10,007,788} & \textbf{1126} & \textbf{8880} & \textbf{0.0000} & \textbf{0.9042} \\
        \bottomrule
    \end{tabular}
\end{table}

\clearpage
\newpage
\bibliographystyle{abbrvnat}
\bibliography{ref_slads}
    
\end{document}